\documentclass[runningheads]{llncs}
\usepackage{graphicx}
\usepackage{tikz}

\usepackage{comment}
\usepackage{amsmath,amssymb} 
\usepackage{color}
\usepackage[font=small,labelfont=bf]{caption}
\usepackage[accsupp]{axessibility}  

\usepackage{booktabs}
\usepackage[normalem]{ulem}
\useunder{\uline}{\ul}{}
\usepackage{longtable}
\usepackage{multirow}

\newcommand{\eg}{\emph{e.g.},\ }
\newcommand{\ie}{\emph{i.e.},\ }
\usepackage[nodisplayskipstretch]{setspace}
\usepackage{cite}
\usepackage{xcolor}

\usepackage[pagebackref=true,breaklinks=true,letterpaper=true,colorlinks,citecolor=blue,bookmarks=false]{hyperref}
             
\begin{document}

\pagestyle{headings}
\mainmatter
\makeatletter
\def\blfootnote{\gdef\@thefnmark{}\@footnotetext}
\makeatother
\def\oursanchor{STARS}
\def\oursmodel{IE-STGCN}
\def\ours{STARS w/ IE-STGCN}
\makeatletter
\newcommand*{\rom}[1]{\expandafter\@slowromancap\romannumeral #1@}
\makeatother
\title{Diverse Human Motion Prediction Guided by Multi-Level Spatial-Temporal Anchors} 

\titlerunning{Diverse Human Motion Prediction Guided by Spatial-Temporal Anchors}

\author{Sirui Xu\index{Xu, Sirui} \and
Yu-Xiong Wang$^*$\index{Wang, Yu-Xiong} \and
Liang-Yan Gui$^*$\index{Gui, Liangyan}}

\authorrunning{S. Xu \and Y.-X. Wang \and L.-Y. Gui}

\institute{University of Illinois at Urbana-Champaign\\
\email{\{siruixu2, yxw, lgui\}@illinois.edu}\\
\url{https://sirui-xu.github.io/STARS/}}
\maketitle
\def\thefootnote{*}\footnotetext{Yu-Xiong Wang and Liang-Yan Gui contributed equally to this work.}\def\thefootnote{\arabic{footnote}}

\begin{abstract}
Predicting diverse human motions given a sequence of historical poses has received increasing attention. Despite rapid progress, existing work captures the multi-modal nature of human motions primarily through likelihood-based sampling, where the mode collapse has been widely observed. In this paper, we propose a simple yet effective approach that disentangles randomly sampled codes with a {\em deterministic learnable component named anchors} to promote sample precision and diversity. Anchors are further factorized into spatial anchors and temporal anchors, which provide attractively {\em interpretable} control over spatial-temporal disparity. In principle, our spatial-temporal anchor-based sampling (\oursanchor{}) can be applied to different motion predictors. Here we propose an interaction-enhanced spatial-temporal graph convolutional network (\oursmodel{}) that encodes prior knowledge of human motions (\eg spatial locality), and incorporate the anchors into it. Extensive experiments demonstrate that our approach outperforms state of the art in both stochastic and deterministic prediction, suggesting it as a {\em unified} framework for modeling human motions. Our code and pretrained models are available at \url{https://github.com/Sirui-Xu/STARS}.

\end{abstract}
\section{Introduction}
\label{sec:intro}
\begin{figure}
\centering
\includegraphics[width=0.92\textwidth]{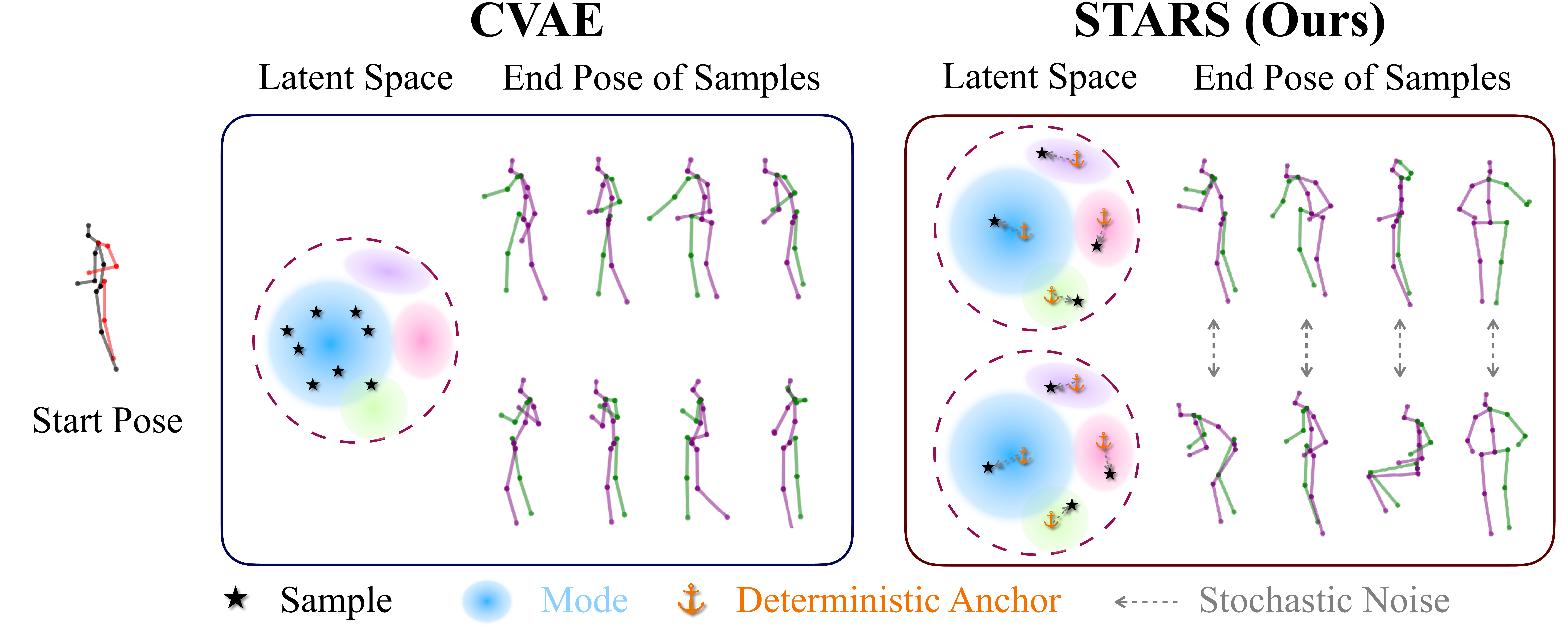}
\caption{
Our Spatial-Temporal AnchoR-based Sampling (\oursanchor) is able to \textit{capture multiple modes}, thus facilitating diverse human motion prediction. {\bf{Left:}} with the traditional generative models such as conditional variational autoencoders (CVAEs), the predicted motions are often concentrated in the major mode with less diversity (illustrated with 8 samples). {\bf{Right:}} \oursanchor~is able to cover more modes, where motions in the same mode have similar characteristics but vary widely across modes. Here, we use 4 anchors to pinpoint different modes. With each anchor, we sample noise and generate 2 similar motions with slight variation in each mode}
\vspace{-1.0em}
\label{fig:diverse_icon}
\end{figure}
Predicting the evolution of the surrounding physical world over time is an essential aspect of human intelligence. For example, in a seamless interaction, a robot is supposed to have some notion of how people move or act in the near future, conditioned on a series of historical movements. Human motion prediction has thus been widely used in computer vision and robotics, such as autonomous driving~\cite{Paden2016ASO}, character animation~\cite{Starke2021NeuralAL}, robot navigation~\cite{rudenko2018joint}, motion tracking~\cite{luber2010people}, and human-robot interaction~\cite{butepage2018anticipating,koppula2016anticipating,Koppula2013AnticipatingHA,lasota2017multiple}. Owing to deep learning techniques, there has been significant progress over the past few years in modeling and predicting motions. Despite notable successes, forecasting human motions, especially over longer time horizons (\ie up to several seconds), is fundamentally challenging, because of the difficulty of modeling multi-modal motion dynamics and uncertainty of human conscious movements. Learning such an uncertainty can, for example, help reduce the search space in motion tracking problems. 

As a powerful tool, deep generative models are thus introduced for this purpose, where random codes from a prior distribution are employed to capture the multi-modal distribution of future human motions.
However, current motion capture datasets are typically constructed in a way that there is \textit{only a single ground truth future sequence} for every single historical sequence~\cite{h36m_pami,Sigal:IJCV:10b}, which makes it difficult for generators to model the underlying multi-modal densities of future motion distribution. Indeed, in practice, generators tend to ignore differences in random codes and simply produce similar predictions. This is known as \textit{mode collapse} -- the samples are concentrated in the major mode, as depicted with a representative example in Fig.~\ref{fig:diverse_icon}, which has been widely observed~\cite{yuan2020dlow}. Recent work has alleviated this problem by explicitly promoting diversity in sampling using post-hoc diversity mappings~\cite{yuan2020dlow}, or through sequentially generating different body parts~\cite{mao2021generating} to achieve combinatorial diversity. These techniques, however, induce additional modeling complexity, without guaranteeing that the diversity modeling accurately covers multiple plausible modes of human motions.

To this end, we propose a simple yet effective strategy --  Multi-Level \textbf{S}patial-\textbf{T}emporal \textbf{A}ncho\textbf{R}-Based \textbf{S}ampling (\oursanchor) -- with the {\em key insight} that future motions are not completely random or independent of each other; they share some deterministic properties in line with physical laws and human body constraints, and continue trends of historical movements. For example, we may expect changes in velocity or direction to be shared deterministically among some future motions, whereas they might differ in magnitude stochastically. Based on this observation, we disentangle latent codes in the generative model into a \textit{stochastic} component (noise) and a \textit{deterministic learnable} component named {\em anchors}. With this disentanglement, the diversity of predictions is jointly affected by random noise as well as anchors that are learned to be specialized for certain modes of future motion. On the contrary, the diversity of traditional generative models is determined by solely independent noise, as depicted in Fig.~\ref{fig:diverse_icon}.
Now, on the one hand, random noise only accounts for modeling the uncertainty {\em within} the mode identified by the anchor, which reduces the burden of having to model the entire future diversity. On the other hand, the model can better capture deterministic states of multiple modes by directly optimizing the anchors, thereby reducing the modeling complexity.

Naturally, human motions exhibit variation in the spatial and temporal domains, and these two types of variation are comparatively independent. Inspired by this, we propose a further decomposition to {\em factorize anchors into spatial and temporal anchors}. Specifically, our designed spatial anchors capture future motion variation at the spatial level, but remain constant at the temporal level, and vice versa. Another appealing property of our approach is that, by introducing straightforward linear interpolation of spatial-temporal anchors, we achieve flexible and seamless control over the predictions (Fig.~\ref{fig:vis_control} and Fig.~\ref{fig:vis_control_interpolate}). Unlike low-level controls that combine motions of different body parts~\cite{mao2021generating,yuan2020dlow}, our work enables manipulation of future motions in the {\em native} \textit{space} and \textit{time}, which is an under-explored problem. Additionally, we propose a multi-level mechanism for spatial-temporal anchors to capture multi-scale modes of future motions.

As a key advantage, spatial-temporal anchors are compatible with any motion predictor. Here, we introduce an \textbf{I}nteraction-\textbf{E}nhanced \textbf{S}patial-\textbf{T}emporal \textbf{G}raph \textbf{C}ovolutional \textbf{N}etwork (\oursmodel{}). This model encodes the spatial locality of human motion and achieves state-of-the-art performance as a motion predictor.

\textbf{Our contributions} can be summarized as follows. (1) We propose a novel anchor-based generative model that formulates sampling as learning deterministic anchors with likelihood sampling to better capture the multiple modes of human motions. (2) We propose a multi-level spatial-temporal decomposition of anchors for interpretable control over future motions. (3) We develop a spatial-temporal graph neural network with interaction enhancement to incorporate our anchor-based sampling. (4) We demonstrate that our approach, as a {\em unified} framework for modeling human motions, significantly outperforms state-of-the-art models in both diverse and deterministic human motion prediction.
\section{Related Work}
\label{sec:related}
\noindent{\bf Deterministic Motion Prediction.}
Existing work on deterministic human motion forecasting predicts a single future motion based on a sequence of past poses~\cite{Aksan_2019_ICCV,Btepage2017DeepRL,Li_2018_CVPR_2,ldrgcn_cvpr20,10.1145/3474085.3475630}, or video frames~\cite{Chao_2017_CVPR,Zhang_2019_ICCV,yuan2019ego}, or under the constraints of the scene context~\cite{Corona_2020_CVPR,caoHMP2020,hassan_samp_2021}, by using recurrent neural networks (RNNs)~\cite{sut11}, temporal convolutional networks (TCNs)~\cite{bai2018empirical}, and graph neural networks (GNNs)~\cite{kipf2017semi} for sequence modeling.
Common early trends involve the use of RNNs~\cite{Jain_2016_CVPR,Wang_2019_ICCV,Gui_2018_ECCV,gui2018few,gui2018teaching}, which are limited in long-term temporal encoding due to error accumulation~\cite{Fragkiadaki,Martinez_2017_CVPR} and training difficulty~\cite{Pascanu2013OnTD}. Some recent attempts exploit GNNs~\cite{Dang_2021_ICCV,wei2020his} to encode poses from the spatial level, but such work still relies on RNNs~\cite{Li_2020_CVPR}, CNNs~\cite{Li_2018_CVPR,Cui_2021_CVPR,Lebailly_2020_ACCV}, or feed-forward networks~\cite{wei2019motion} for temporal modeling. Recently, spatial-temporal graph convolutional networks (STGCNs)~\cite{yu2018spatio,Sofianos_2021_ICCV,yan2021dmsgcn} have been proposed to jointly encode spatial and temporal correlations with spatial-temporal graphs. Continuing this effort, we propose IE-STGCN, which additionally encodes inductive biases such as spatial locality into STGCNs.

\noindent{\bf Stochastic Motion Prediction.}
Stochastic human motion prediction is an emerging trend with the development of deep generative models such as variational autoencoders (VAEs)~\cite{kingma2014autoencoding}, generative adversarial networks (GANs)~\cite{NIPS2014_5ca3e9b1}, and normalizing flows (NFs)~\cite{JimenezRezende2015VariationalIW}. Most existing work~\cite{Walker2017ThePK,lin2018human,Barsoum2018HPGANP3,Hernandez_2019_ICCV,article,yan2018mt,Aliakbarian_2020_CVPR,yuan2020residual} produces various predictions from a set of codes independently sampled from a given distribution. As depicted in DLow~\cite{yuan2020dlow}, such likelihood-based sampling cannot produce enough diversity, as many samples are merely perturbations in the major mode. To overcome the issue, DLow employs a two-stage framework, using post-hoc mappings to shape and diversify the latent samples. GSPS~\cite{mao2021generating} generates different body parts in a sequential manner to achieve combinatorial diversity. Nevertheless, their explicit promotion of diversity induces additional complexity but does not directly enhance multi-mode capture. We introduce anchors that are comparatively easy to optimize, locate deterministic components of motion modes, and impose sample diversity.

\noindent{\bf Controllable Motion Prediction.}
Controllable motion prediction has been explored in computer graphics for the generation of virtual characters~\cite{Holden2017PhasefunctionedNN,Holden2016ADL,ling2020character}. In the task of human motion prediction, DLow~\cite{yuan2020dlow} and GSPS~\cite{mao2021generating} control the predicted motion by separating upper and lower body parts, fixing one part while controlling the other. In this paper, through the use of spatial-temporal anchors, we propose different but more natural controllability in native space and time. By varying and interpolating the spatial and temporal anchors, we achieve high-level control over the spatial and temporal variation, respectively.

\noindent{\bf Learnable Anchors.}
Our anchor-based sampling, \ie sampling with deterministic learnable codes, is inspired by work on leveraging predefined primitives and learnable codes for applications such as trajectory prediction~\cite{chai2019multipath,cui2019icra,Kothari2021InterpretableSA,liu2021multimodal,Phan-Minh_2020_CVPR}, object detection~\cite{Liu2016SSDSS,carion2020endtoend}, human pose estimation~\cite{Yang2011ArticulatedPE}, and video representation learning~\cite{Han20}. Anchors usually refer to the {\em hypothesis} of predictions, such as box candidates with different shapes and locations in object detection~\cite{Liu2016SSDSS}. In a similar spirit, anchors in the context of human motion prediction indicate assumptions about future movements. The difference is that the anchors here are {\em not hand-crafted or predefined} primitives; instead, they are latent codes learned from the data. In the meantime, we endow anchors with explainability \ie to describe the multi-level spatial-temporal variation of future motions.
\begin{figure}
\centering
\includegraphics[width=0.95\textwidth]{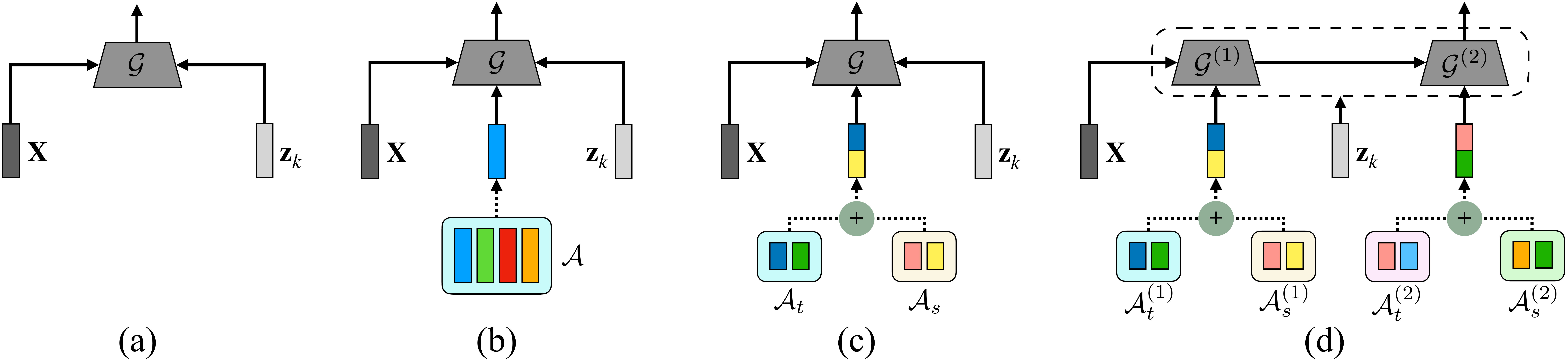}
\caption{\textbf{Comparison of generative models without and with anchor-based sampling.} Anchors and network parameters are jointly optimized. (a) Conventional generative model with only stochastic noise; (b) Generative model with deterministic anchor process: an anchor with Gaussian noise corresponds to a prediction; (c) Spatial-temporal compositional anchors: any pair of combined spatial and temporal anchors corresponds to a prediction; (d) Multi-level spatial-temporal anchors: anchors at different levels are combined for encoding multi-scale modes}
\label{fig:diverse}
\end{figure}

\section{Methodology}
\label{sec:method}

\noindent{\bf Problem Formulation.}
We denote the input motion sequence of length $T_h$ as $\mathbf X=[\mathbf x_1, \mathbf x_2,\ldots,\mathbf x_{T_h}]^T$,
where the 3D coordinates of $V$ joints are used to describe each pose $\mathbf x_i \in \mathbb{R}^{{V}\times C^{(0)}}$. 
Here, we have $C^{(0)} = 3$. 
And $K$ output sequences of length $T_p$ are denoted as $\mathbf{\widehat{Y}}_1, \mathbf{\widehat{Y}}_2, \ldots, \mathbf{\widehat{Y}}_K$. We have access to a {\em{single}} ground truth future motion of length $T_p$ as $\mathbf{Y}$.
{\it{Our objectives}} are: (1) one of the $K$ predictions is as close to the ground truth as possible; and (2) the $K$ sequences are as diverse as possible, yet represent realistic future motions.

In this section, we first briefly review deep generative models, describe how they draw samples to generate multiple futures, and discuss their limitations (Sec.~\ref{sec:shmp}). We then detail our insights on \oursanchor{} including anchor-based sampling and multi-level spatial-temporal anchors (Sec.~\ref{sec:shmp} and Fig.~\ref{fig:diverse}). To model human motion, we design an \oursmodel{} and incorporate our spatial-temporal anchors into it (Sec.~\ref{sec:network}), as illustrated in Fig.~\ref{fig:deterministic2}.

\subsection{Multi-Level Spatial-Temporal Anchor-Based Sampling}\label{sec:shmp}

\noindent{\bf Preliminaries: Deep Generative Models.}
There is a large body of work on the generation of multiple hypotheses with deep generative models, most of which learn a parametric probability distribution function explicitly or implicitly. Let $p(\mathbf{Y}|\mathbf{X})$ denote the distribution of the future human motion $\mathbf{Y}$ conditioned on the past sequence $\mathbf{X}$. With a latent variable $\mathbf z \in \mathcal{Z}$, the distribution can be reparameterized as $p(\mathbf{Y}|\mathbf{X}) = \int p(\mathbf{Y}|\mathbf{X},\mathbf z)p(\mathbf z) \mathrm{d} \mathbf z$, where $p(\mathbf z)$ is often a Gaussian prior distribution. To generate a future motion sequence $\mathbf{\widehat Y}$, $\mathbf{z}$ is drawn from the given distribution $p(\mathbf z)$, and then a deterministic generator $\mathcal{G}:\mathcal{Z}\times \mathcal{X} \rightarrow \mathcal{Y}$ is used for mapping, as illustrated in Fig.~\ref{fig:diverse}{\color{red}{(a)}}: 
\begin{align}
    \mathbf{z} \sim p(\mathbf{z}), \ \mathbf{\widehat Y} = \mathcal{G}(\mathbf{z}, \mathbf{X}),
\end{align}
where $\mathcal{G}$ is a deep neural network parameterized by $\theta$. The goal of generative modeling is to make the distribution $p_{\theta}(\mathbf{\widehat Y}|\mathbf{X})$ derived from the generator $\mathcal{G}$ close to the actual distribution $p(\mathbf{Y}|\mathbf{X})$. 

To generate $K$ diverse motion predictions, traditional approaches first independently sample a set of latent codes $Z = \{\mathbf{z}_1, \ldots, \mathbf{z}_K\}$ from a prior distribution $p(\mathbf{z})$.
Although in theory, generative models are capable of covering different modes, they are not guaranteed to locate all modes precisely, and mode collapse has been widely observed \cite{yuan2020dlow,Yuan2020Diverse}.

\noindent{\bf Anchor-Based Sampling.}
To address this problem, we propose a simple, yet effective sampling strategy. Our intuition is that the diversity in future motions could be characterized by: (1) deterministic component -- across different actions performed by different subjects, there exist correlated or shareable changes in velocity, direction, movement patterns, etc., which naturally emerge and can be directly learned from data; and (2) stochastic component -- given an action carried out by a subject, the magnitude of the changes exists, which is stochastic.

Therefore, we disentangle the code in the latent space of the generative model into a {\em{stochastic}} component sampled from $p(\mathbf{z})$, and a {\em{deterministic}} component represented by a set of $K$ {\em{learnable parameters}} called {\em anchors} $\mathcal{A} = \{\mathbf a_k\}_{k=1}^K$. Deterministic anchors are expected to identify as many modes as possible, which is achieved through a carefully designed optimization, while stochastic noise further specifies motion variation within certain modes. With this latent code disentanglement, we denote the new multi-modal distribution as
\begin{align}
    p_{\theta}(\mathbf{\widehat Y}|\mathbf{X}, \mathcal{A}) = \frac{1}{K}\sum_{k=1}^K \int p_{\theta}(\mathbf{\widehat Y}|\mathbf{X}, \mathbf z,\mathbf a_k)p(\mathbf z) \mathrm{d} \mathbf z.
\end{align}

Consequently, as illustrated in Fig.~\ref{fig:diverse}{\color{red}{(b)}}, suppose that we select the $k$-th learned anchor $\mathbf a_k \in \mathcal{A}$, along with the randomly sampled noise $\mathbf z \in Z$, we can generate the prediction $\mathbf{\widehat{Y}}_k$ as,
\begin{align}
    \mathbf{z} \sim p(\mathbf z),\ \mathbf{\widehat{Y}}_k = \mathcal{G}(\mathbf{a}_k, \mathbf{z}, \mathbf{X}).
\end{align}

\begin{figure}[t]
\centering
\includegraphics[width=\textwidth]{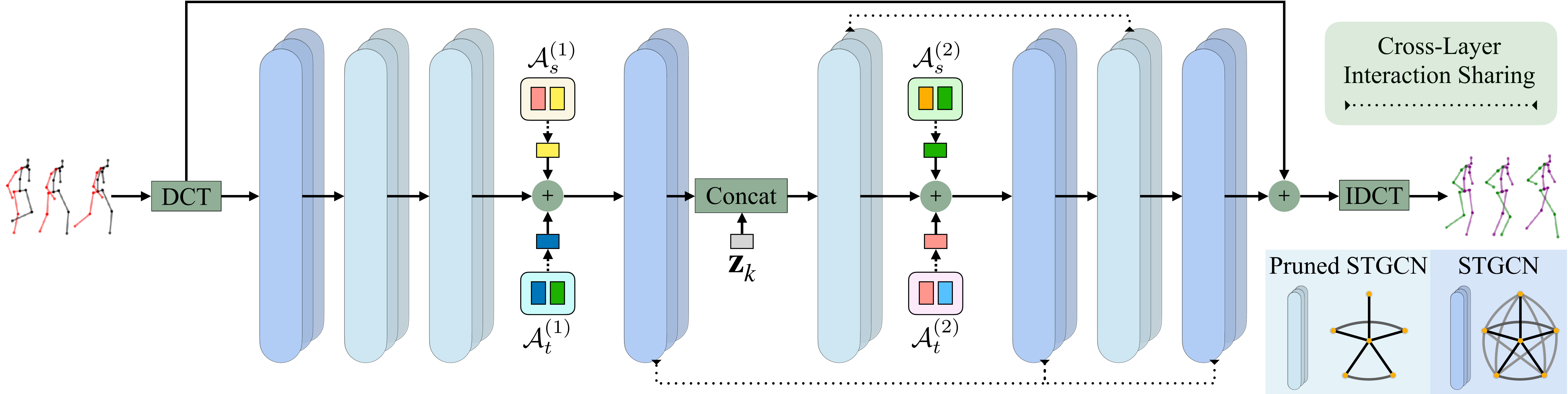}
\caption{\textbf{Overview of our \ours{} framework.} We combine multi-level spatial-temporal anchors, the sampled noise, with the backbone \oursmodel. To generate one of the predictions given a past motion, we draw noise $\mathbf z_k$, and add the selected spatial-temporal anchors to the latent feature at each level}
\label{fig:deterministic2}
\end{figure}

We can produce a total of $K$ predictions if using each anchor once, though all anchors are not limited to being used or used only once. To incorporate anchors into the network, we find it effective to make simple additions between selected anchors and latent features, as shown in Fig.~\ref{fig:deterministic2}.

\noindent{\bf Spatial-Temporal Compositional Anchors.}\label{sec:mmst} We observe that the diversity of future motions can be roughly divided into two types, namely {\em{spatial variation}} and {\em{temporal variation}}, which are relatively independent. This sheds light on a feasible further decomposition of the $K$ anchors into two types of learnable codes: spatial anchors $\mathcal{A}_s = \{\mathbf a^s_i\}_{i=1}^{K_s}$ and temporal anchors $\mathcal{A}_t = \{\mathbf a^t_j\}_{j=1}^{K_t}$, where $K = K_s \times K_t$. 
With this decomposition, we still can yield a total of $K_s \times K_t$ compositional anchors through \textit{each pair of spatial-temporal anchors}.
Note that the temporal anchors here, in fact, control the frequency variation of future motion sequences, since our temporal features are in the frequency domain, as we will demonstrate in Sec.~\ref{sec:network}. 
To be more specific, conceptually, all spatial anchors are set to be identical in the temporal dimension but characterize the variation of motion in the spatial dimension, taking control of the movement trends and directions. Meanwhile, all temporal anchors remain unchanged in the spatial dimension but differ in the temporal dimension, producing disparities in frequency to affect the movement speed. 

To produce $\mathbf{\widehat{Y}}_k$, as depicted in Fig.~\ref{fig:diverse}{\color{red}{(c)}}, we sample $\mathbf z$ and select $i$-th spatial anchor $\mathbf a_i^s$ and $j$-th temporal anchor $\mathbf a_j^t$, 
\begin{align}
    \mathbf{z} \sim p(\mathbf z),\ \mathbf{\widehat{Y}}_k = \mathcal{G}(\mathbf a_i^s + \mathbf a_j^t, \mathbf{z}, \mathbf{X}),
\end{align}
where $\mathbf a_i^s + \mathbf a_j^t$ is a spatial-temporal compositional anchor corresponding to an original anchor $\mathbf a_k$. Furthermore, motion control over spatial and temporal variation can be customized through these spatial-temporal anchors. For example, we can produce future motions with similar trends by fixing the spatial anchors while varying or interpolating the temporal anchors, as shown in Sec.~\ref{sec:qual}. 

\noindent{\bf Multi-Level Spatial-Temporal Anchors.}
To further learn and capture multi-scale modes of future motions, we propose a multi-level mechanism to extend the spatial-temporal anchors.
As an illustration, Fig.~\ref{fig:diverse}{\color{red}{(d)}} shows a simple two-level case for this design. We introduce two different spatial-temporal anchor sets, $\{\mathcal{A}_t^{(1)},\mathcal{A}_s^{(1)}\}$ and $\{\mathcal{A}_t^{(2)},\mathcal{A}_s^{(2)}\}$, and assign them sequentially to different network parts $\mathcal{G}^{(1)},\mathcal{G}^{(2)}$. Suppose $(i, j)$ is a spatial-temporal index corresponding to the 1D index $k$, we can generate $\mathbf{\widehat{Y}}_k$ through a two-level process as
\begin{align}
    \mathbf{z} \sim p(\mathbf z),\ \mathbf{\widehat{Y}}_k = \mathcal{G}^{(2)}(\mathbf a_i^{s_{2}} + \mathbf a_j^{t_{2}}, \mathbf{z}, \mathcal{G}^{(1)}(\mathbf a_i^{s_{1}} + \mathbf a_j^{t_{1}}, \mathbf{X})),
\end{align}
where $a_i^{s_{1}} \in \mathcal{A}_s^{(1)}, a_j^{t_{1}} \in \mathcal{A}_t^{(1)}, a_i^{s_{2}} \in \mathcal{A}_s^{(2)}, a_j^{t_{2}} \in \mathcal{A}_t^{(2)}$. As a principled way, anchors can be applied at more levels to encode richer assumptions about future motions.
 
\noindent{\bf Training.}
During training, the model uses \textit{each spatial-temporal anchor} explicitly to generate $K$ future motions for each past motion sequence. The loss functions are mostly adopted as proposed in~\cite{mao2021generating}, which we summarize into three categories: (1) reconstruction losses that optimize \textit{the best predictions} under different definitions among $K$ generated motions, and thus optimize anchors to their own nearest modes; (2) a diversity-promoting loss that explicitly promotes pairwise distances in predictions, avoiding that anchors collapse to the same; and (3) motion constraint losses that encourage output movements to be realistic. All anchors are directly learned from the data via gradient descent.
In the forward pass, we explicitly take {\em every} anchor $\mathbf{a}_i \in \mathcal{A}=\{\mathbf{a}_k\}_{k=1}^K$ as an additional input to the network and produce a total of $K$ outputs. In the backward pass, each anchor is optimized \textit{separately} based on its corresponding outputs and losses, while the backbone network is updated based on the \textit{fused} losses from all outputs. This separate backward pass is automatically done via PyTorch~\cite{paszke2019pytorch}. Please refer to Sec.~\ref{sec:shmp_supp} of the supplementary material for more details.

\subsection{Interaction-Enhanced Spatial-Temporal Graph Convolutional Network}\label{sec:network}
In principle, our proposed anchor-based sampling permits flexible network architectures. Here, to incorporate our multi-level spatial-temporal anchors, we naturally represent motion sequences as spatial-temporal graphs (to be precise,  spatial-frequency graphs), instead of the widely used spatial graphs~\cite{wei2019motion,mao2021generating}.  Our approach builds upon the Discrete Cosine Transform (DCT)~\cite{wei2019motion,mao2021generating} to transform the motion into the frequency domain. Specifically, given a past motion $\mathbf X_{1:T_h} \in \mathbb{R}^{T_h \times V \times C^{(0)}}$, where each pose has $V$ joints, we first replicate the last pose $T_p$ times to get $\mathbf {X}_{1:T_h+T_p} = [\mathbf x_1, \mathbf x_2,\ldots,\mathbf x_{T_h}, \mathbf x_{T_h}, \ldots,\mathbf x_{T_h}]^T$. With the predefined $M$ basis $\mathbf{C} \in \mathbb{R}^{ M\times (T_h+T_p)}$ for DCT, the motion is transformed as
\begin{align}
    \widetilde{\mathbf {X}} = \mathbf{C}\mathbf {X}_{1:T_h+T_p}.
\end{align}

We formulate $\widetilde{\mathbf {X}} \in \mathbb{R}^{M \times V \times C^{(0)}}$ in the $0$-th layer and latent features in any $l$-th graph layer as spatial-temporal graphs $(\mathcal{V}^{(l)}, \mathcal{E}^{(l)})$ with $M \times V$ nodes. We specify the node $i$ by the 2D index $(f_i, v_i)$ for the joint $v_i$ with frequency $f_i$ component. The edge $(i, j) \in \mathcal{E}^{(l)}$ associated with the interaction between node $i$ and node $j$ is represented by $\mathbf {Adj}^{(l)}[i][j]$, where the adjacency matrix $\mathbf {Adj}^{(l)} \in \mathbb{R}^{M V\times M V}$ is learnable. We bottleneck spatial-temporal interactions as~\cite{Sofianos_2021_ICCV}, by factorizing the adjacency matrix into the product of low-rank spatial and temporal matrices $\mathbf {Adj}^{(l)} = \mathbf {Adj}^{(l)}_{s} \mathbf {Adj}^{(l)}_{f}$. The spatial adjacency matrix $\mathbf {Adj}^{(l)}_{s} \in \mathbb{R}^{M V\times M V}$ connects only nodes with the same frequency. And the frequency adjacency matrix $\mathbf {Adj}^{(l)}_{f} \in \mathbb{R}^{M V\times M V}$ is merely responsible for the interaction between the nodes that represent the same joint.

The spatial-temporal graph can be conveniently encoded by a graph convolutional network (GCN). Given a set of trainable weights $\mathbf W^{(l)} \in \mathbb{R}^{C^{(l)}\times C^{(l+1)}}$ and activation function $\sigma(\cdot)$, such as ReLU, a spatial-temporal graph convolutional layer projects the input from $C^{(l)}$ to $C^{(l+1)}$ dimensions by
\begin{align}\label{eq:1}
    \mathbf{H}^{(l+1)}_k = \sigma(\mathbf {Adj}^{(l)}\mathbf{H}^{(l)}_k\mathbf W^{(l)})=\sigma(\mathbf {Adj}_{s}^{(l)}\mathbf {Adj}_{f}^{(l)}\mathbf{H}^{(l)}_k\mathbf W^{(l)}),
\end{align}
where $\mathbf{H}^{(l)}_k \in \mathbb{R}^{M V \times C^{(l)}}$ denotes the latent feature of the prediction $\mathbf{\widehat{Y}}_k$ at $l$-th layer. 
The backbone consists of multiple graph convolutional layers. After generating predicted DCT coefficients $\widetilde{\mathbf {Y}}_k \in \mathbb{R}^{M \times V \times C^{(L)}}$ reshaped from $\mathbf{H}^{(L)}_k$, where $C^{(L)} = 3$, we recover $\mathbf{\widehat{Y}}_k$ via Inverse DCT (IDCT) as
\begin{align}
    \mathbf {\widehat Y}_k = (\mathbf{C^\mathsf{T}}\widetilde{\mathbf {Y}}_k)_{T_h+1:T_h+T_p},
\end{align}
where the last $T_p$ frames of the recovered sequence represent future poses.

Conceptually, interactions between spatial-temporal nodes should be relatively invariant across layers, and different interactions should not be equally important. For example, we expect constraints and dependencies between ``left arm'' and ``left forearm,'' while the movements of ``head'' and ``left forearm'' are relatively independent.
We consider it redundant to construct a {\em complete} spatial-temporal graph for each layer {\em independently}. Therefore, we introduce cross-layer interaction sharing to share parameters between graphs in different layers, and spatial interaction pruning to prune the complete graph.

\noindent{\bf Cross-Layer Interaction Sharing.} 
Much care has been taken to employ learnable interactions between spatial nodes across all graph layers~\cite{Sofianos_2021_ICCV,wei2019motion,yan2021dmsgcn}.
We consider the spatial relationship to be relatively unchanged. Empirically, we find that sharing the adjacency matrix at intervals of one layer is effective. As shown in Fig.~\ref{fig:deterministic2}, we set $\mathbf {Adj}_s^{(4)} = \mathbf {Adj}_s^{(6)} = \mathbf {Adj}_s^{(8)}$ and $\mathbf {Adj}_s^{(5)} = \mathbf {Adj}_s^{(7)}$.

\noindent{\bf Spatial Interaction Pruning.}
To emphasize the physical relationships and constraints between spatial joints, we prune the spatial connections $\mathbf {\widehat{Adj}}_s^{(l)} = \mathbf M_s \odot \mathbf {Adj}_s^{(l)}$ in each graph layer $l$ using a predefined mask $\mathbf M_s$, where $\odot$ is an element-wise product.
Inspired by~\cite{Liu_2021_ICCV}, we emphasize spatial locality based on skeletal connections and mirror symmetry tendencies. We denote our proposed predefined mask matrix as
\begin{align}
    \mathbf M_s[i][j] = \begin{cases} 1, & v_i \mbox{ and } v_j \mbox{ are physically connected, }f_i=f_j \\
    1, & v_i \mbox{ and } v_j \mbox{ are mirror-symmetric, }f_i=f_j  \\ 0, & \mbox{otherwise}. \end{cases}
\end{align}

Finally, our architecture consists of four original \textbf{STGCNs} without spatial pruning and four \textbf{Pruned STGCNs}, as illustrated in Fig.~\ref{fig:deterministic2}. Please refer to Sec.~\ref{sec:shmp_supp} of the supplementary material for more information on the architecture.

\section{Experiments}\label{sec:exp}
\subsection{Experimental Setup for Diverse Prediction}\label{sec: metrics}

\noindent\textbf{Datasets.}
We perform an evaluation on two motion capture datasets, Human3.6M \cite{h36m_pami} and HumanEva-I~\cite{Sigal:IJCV:10b}. Human3.6M consists of 11 subjects and 3.6 million frames at 50 Hz. 
Following~\cite{mao2021generating,yuan2020dlow}, we use a 17-joint skeleton representation and train our model to predict 100 future frames given 25 past frames without global translation. We train on five subjects (S1, S5, S6, S7, and S8) and test on two subjects (S9 and S11). 
HumanEva-I contains 3 subjects recorded at 60 Hz. 
Following~\cite{mao2021generating,yuan2020dlow}, the pose is represented by 15 joints. We use the official train/test split~\cite{Sigal:IJCV:10b}. The model forecasts 60 future frames given 15 past frames.

\noindent\textbf{Metrics.}
For a fair comparison, we measure the diversity and accuracy of the predictions according to the evaluation metrics in~\cite{Yuan2020Diverse,yuan2020dlow,mao2021generating,Aliakbarian_2020_CVPR}. (1) \textbf{Average Pairwise Distance (APD)}: average $\ell_2$ distance between all prediction pairs, defined as $\frac{1}{K(K-1)} \sum_{i=1}^K \sum_{j\neq i}^K \|\widehat{\mathbf{Y}}_i - \widehat{\mathbf{Y}}_j\|_2$. (2) \textbf{Average Displacement Error (ADE)}: average $\ell_2$ distance over the time between the ground truth and the closest prediction, computed as $\frac{1}{T_p}\min_k \|\widehat{\mathbf{Y}}_k - \mathbf{Y}\|_2$. (3) \textbf{Final Displacement Error (FDE)}: $\ell_2$ distance of the last frame between the ground truth and the closest prediction, defined as $\min_k\|\widehat{\mathbf{Y}}_k[T_p] - \mathbf{Y}[T_p]\|_2$. To measure the ability to produce multi-modal predictions, we also report multi-modal versions of ADE and FDE. We define the multi-modal ground truth~\cite{Yuan2020Diverse} as $\{\mathbf Y_n\}_{n=1}^N$, which is clustered based on historical pose distances, representing possible multi-modal future motions. The detail of multi-modal ground truth is in the supplementary material. (4) \textbf{Multi-Modal ADE (MMADE)}: the average displacement error between the predictions and the multi-modal ground truth, denoted as $\frac{1}{NT_p}\sum_{n=1}^N\min_k\|\widehat{\mathbf{Y}}_k - \mathbf{Y}_n\|_2$.  (5) \textbf{Multi-Modal FDE (MMFDE)}: the final displacement error between the predictions and the multi-modal ground truth, denoted as $\frac{1}{N}\sum_{n=1}^N\min_k\|\widehat{\mathbf{Y}}_k[T_p] - \mathbf{Y}_n[T_p]\|_2$. All metrics here are in \textit{meters}.

\noindent\textbf{Baselines.}
To evaluate our stochastic motion prediction method, we consider two types of baselines: (1) Stochastic methods, including CVAE-based methods, \textbf{Pose-Knows}~\cite{Walker2017ThePK} and \textbf{MT-VAE}~\cite{yan2018mt}, as well as CGAN-based methods, \textbf{HP-GAN}~\cite{Barsoum2018HPGANP3}; (2) Diversity promoting methods, including \textbf{Best-of-Many}~\cite{apratim18cvpr2}, \textbf{GMVAE}~\cite{Dilokthanakul2016DeepUC}, \textbf{DeLiGAN}~\cite{Gurumurthy_2017_CVPR}, \textbf{DSF}~\cite{Yuan2020Diverse}, \textbf{DLow}~\cite{yuan2020dlow}, \textbf{MOJO}~\cite{Zhang_2021_CVPR}, and \textbf{GSPS}~\cite{mao2021generating}.

\noindent\textbf{Implementation Details.}
The backbone consists of 8 GCN layers. We perform spatial pruning on 4 GCN layers (denoted as `Pruned’). The remaining 4 layers are not pruned. In each layer, we use batch normalization~\cite{pmlr-v37-ioffe15} and residual connections.
We add $K$ spatial-temporal compositional anchors at layers $4$ and $6$, and random sampling at layer $5$. Here, $K=50$ unless otherwise specified.
For Human3.6M, the model is trained for 500 epochs, with a batch size of 16 and 5000 training instances per epoch. 
For HumanEva-I, the model is trained for 500 epochs, with a batch size of 16 and 2000 training instances per epoch. 
Additional implementation details are provided in Sec.~\ref{sec:shmp_supp} of the supplementary material.

\subsection{Quantitative Results and Ablation of Diverse Prediction}\label{sec:diverse_ablation}
We compare our method with the baselines in Table \ref{tab:diverse_quan} on Human3.6M and HumanEva-I. We produce one prediction using \textit{each spatial-temporal anchor} for a total of $50$ predictions, which is consistent with the literature~\cite{mao2021generating,yuan2020dlow}. For all metrics, our method consistently outperforms all baselines on both datasets. Methods such as GMVAE~\cite{Dilokthanakul2016DeepUC} and DeLiGAN~\cite{Gurumurthy_2017_CVPR} have relatively low accuracy (ADE, FDE, MMADE, and MMFDE) and diversity (APD), since they still follow a pure random sampling. Methods such as DSF~\cite{Yuan2020Diverse}, DLow~\cite{yuan2020dlow} and GSPS~\cite{mao2021generating} explicitly promote diversity by introducing assumptions in the latent codes or directly in the generation process. Instead, we propose to use anchors to locate diverse modes directly learned from the data, which is more effective.

\begin{table}[t]
	\caption{\textbf{Quantitative results} on Human3.6M and HumanEva-I for $K=50$. Our model significantly outperforms all stochastic prediction baselines on all metrics. The results of baselines are reported from \cite{mao2021generating,yuan2020dlow,Zhang_2021_CVPR}}

	\centering
	\resizebox{\textwidth}{!}{
		\begin{tabular}{cccccccccccc}
				\noalign{\smallskip}
		\noalign{\smallskip}
			\toprule
			\multirow{2}{*}{Method}& \multicolumn{5}{c}{Human3.6M~\cite{h36m_pami}} & & \multicolumn{5}{c}{HumanEva-I~\cite{Sigal:IJCV:10b}} \\ \cmidrule{2-6} \cmidrule{8-12}
			 & APD $\uparrow$ & ADE $\downarrow$ & FDE $\downarrow$ & MMADE $\downarrow$ & MMFDE $\downarrow$ & & APD $\uparrow$ & ADE $\downarrow$ & FDE $\downarrow$ & MMADE $\downarrow$ & MMFDE $\downarrow$ \\ \midrule

			Pose-Knows \cite{Walker2017ThePK}              & 6.723  & 0.461 & 0.560 & 0.522 & 0.569 &  & 2.308 & 0.269 & 0.296 & 0.384 & 0.375 \\
			MT-VAE \cite{yan2018mt}           & 0.403  & 0.457 & 0.595 & 0.716 & 0.883 &  & 0.021 & 0.345 & 0.403 & 0.518 & 0.577 \\
			HP-GAN \cite{Barsoum2018HPGANP3}           & 7.214  & 0.858 & 0.867 & 0.847 & 0.858 &  & 1.139 & 0.772 & 0.749 & 0.776 & 0.769 \\\hline
			BoM \cite{apratim18cvpr2} & 6.265  & 0.448 & 0.533 & 0.514 & 0.544 &  & 2.846 & 0.271 & 0.279 & 0.373 & 0.351 \\
			GMVAE \cite{Dilokthanakul2016DeepUC}            & 6.769  & 0.461 & 0.555 & 0.524 & 0.566 &  & 2.443 & 0.305 & 0.345 & 0.408 & 0.410 \\
			DeLiGAN \cite{Gurumurthy_2017_CVPR}          & 6.509  & 0.483 & 0.534 & 0.520 & 0.545 &  & 2.177 & 0.306 & 0.322 & 0.385 & 0.371 \\
			DSF \cite{Yuan2020Diverse}                    & 9.330  & 0.493 & 0.592 & 0.550 & 0.599 &  & 4.538 & 0.273 & 0.290 & 0.364 & 0.340 \\
			DLow~\cite{yuan2020dlow} & 11.741 & 0.425 & 0.518 & 0.495 & 0.531 &  & 4.855 & 0.251 & 0.268 & 0.362 & 0.339 \\
			MOJO \cite{Zhang_2021_CVPR} & 12.579 & 0.412 & 0.514 & 0.497 & 0.538 & & 4.181 & 0.234 & 0.244 & 0.369 & 0.347 \\
			GSPS \cite{mao2021generating} & 14.757 & 0.389 & 0.496 & 0.476 & 0.525 & &5.825 & 0.233 & 0.244 & 0.343 & 0.331 \\\hline
			\oursanchor{} (Ours) & \textbf{15.884} & \textbf{0.358} & \textbf{0.445} & \textbf{0.442} & \textbf{0.471} & & \textbf{6.031} & \textbf{0.217} & \textbf{0.241} & \textbf{0.328} & \textbf{0.321}\\
			\bottomrule
		\end{tabular}
	}

	\label{tab:diverse_quan}

\end{table}

\begin{table}[t]
	\caption{\textbf{Ablation study} on Human3.6M and HumanEva-I for $K=50$. We compare the following $4$ cases: (\rom{1}) $50$ original anchors; (\rom{2}) $2$ temporal anchors and $25$ spatial anchors; (\rom{3}) $50$ spatial-temporal compositional anchors from $5$ temporal anchors and $10$ spatial anchors; (\rom{4}) $50$ spatial-temporal compositional anchors for both levels}

	\centering
	\resizebox{\textwidth}{!}{
		\begin{tabular}{cccccccccccc}
				\noalign{\smallskip}
		\noalign{\smallskip}
			\toprule
			\multirow{2}{*}{$\#$ of anchors}& \multicolumn{5}{c}{Human3.6M~\cite{h36m_pami}} & & \multicolumn{5}{c}{HumanEva-I~\cite{Sigal:IJCV:10b}} \\ \cmidrule{2-6} \cmidrule{8-12}
			 & APD $\uparrow$ & ADE $\downarrow$ & FDE $\downarrow$ & MMADE $\downarrow$ & MMFDE $\downarrow$ & & APD $\uparrow$ & ADE $\downarrow$ & FDE $\downarrow$ & MMADE $\downarrow$ & MMFDE $\downarrow$ \\ \midrule

			(\rom{1}) 50 & \textbf{16.974} & 0.363 & 0.447 & 0.444 & 0.473 & & \textbf{7.786} & 0.221 & 0.249 & \textbf{0.327} & \textbf{0.321}\\

			(\rom{2}) $2 \times 25$  & 16.303 & 0.356 & \textbf{0.442} & 0.440 & 0.468 & & 5.254 & 0.224 & 0.253 & 0.337 & 0.331\\
			(\rom{3}) $5 \times 10$  & 13.681 & \textbf{0.355} & \textbf{0.442} & \textbf{0.439} & \textbf{0.467} & & 6.199 & 0.226 & 0.255 & 0.334 & 0.330\\
			(\rom{4}) $(5 \times 10) \times 2$ &15.884 & 0.358 & 0.445 & 0.442 & 0.471 & & 6.031 & \textbf{0.217} & \textbf{0.241} & 0.328 & \textbf{0.321}\\
			\bottomrule
		\end{tabular}
	}

	\label{tab:diverse_ablation}

\end{table}

\noindent\textbf{Effectiveness of multi-level spatial-temporal anchors.}
As shown in Table~\ref{tab:diverse_ablation}, compared to not using spatial-temporal decoupling (\rom{1}), using it (\rom{2} and \rom{3}) leads to relatively lower diversity, but facilitates mode capture and results in higher accuracy on Human3.6M.
Applying the multi-level mechanism (\rom{4}) improves diversity, but sacrifices a little accuracy on Human3.6M. On the contrary, we observe improvements in both diversity and accuracy on HumanEva-I. The results suggest that there is an intrinsic trade-off between diversity and accuracy. Higher diversity indicates that the model has a better chance of covering multiple modes. However, when diversity exceeds a certain level, the trade-off between diversity and accuracy becomes noticeable.

\begin{figure}[t]
\centering
\includegraphics[width=\textwidth]{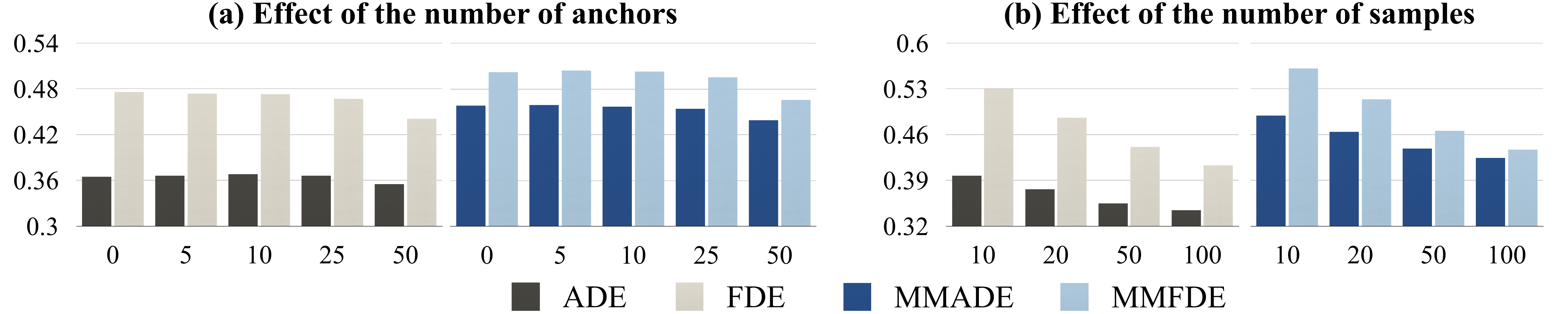}
\caption{\textbf{Ablation study} on Human3.6M. We report ADE, MMADE, FDE, and MMFDE, comparing settings with different numbers of anchors and samples}
\label{fig:quan_num}
\end{figure}
\noindent\textbf{Impact of number of anchors and samples.}
We investigate the effect of two important hyperparameters on the model, \ie the number of anchors and the number of samples. As illustrated in Fig.~\ref{fig:quan_num}{\color{red}{(a)}}, we fix the number of samples to $50$ and compare the results when the number of anchors varies within $0,5,10,25,50$. The results show that more anchors enable the model to better capture the major modes (ADE, FDE) and also other modes (MMADE, MMFDE). In Fig.~\ref{fig:quan_num}{\color{red}{(b)}}, we vary the sample size to $10, 20, 50, 100$ and keep the number of anchors the same as the number of samples. The results show that the larger the number of samples, the easier it is for a sample to approach the ground truth.

\begin{table}[t]
	\caption{\textbf{Ablation study} on Human3.6M for $K=100$. We demonstrate the generalizability of our anchor-based sampling. For a fair comparison, we add single-level anchor-based sampling to GSPS~\cite{mao2021generating} and \oursmodel{}, without changing any other design and without using spatial-temporal decomposition. We observe that our anchor-based sampling mechanism consistently improves diversity and accuracy for both approaches. Meanwhile, our backbone is more lightweight but performs better}
	\centering
	\resizebox{\textwidth}{!}{
		\begin{tabular}{cccccccccccccc}
				\noalign{\smallskip}
		\noalign{\smallskip}
			\toprule
			\multirow{2}{*}{Backbone} & \multirow{2}{*}{Parameter} & & \multicolumn{5}{c}{random sampling} & & \multicolumn{5}{c}{single-level anchor-based sampling (Ours)} \\ \cmidrule{4-8} \cmidrule{10-14}
			 &&& APD $\uparrow$ & ADE $\downarrow$ & FDE $\downarrow$ & MMADE $\downarrow$ & MMFDE $\downarrow$ & & APD $\uparrow$ & ADE $\downarrow$ & FDE $\downarrow$ & MMADE $\downarrow$ & MMFDE $\downarrow$ \\ \midrule

			GSPS \cite{mao2021generating} & 1.30M && 14.751 & 0.369 & 0.462 & 0.455 & 0.491 & & 19.022 & 0.364 & 0.443 & 0.443 & 0.465 \\

			\oursmodel{} (Ours) & 0.29M && 11.701 & 0.352 & 0.445 & 0.443 & 0.476 & & 14.554 & 0.344 & 0.411 & 0.423 & 0.436\\
			\bottomrule
		\end{tabular}
	}

	\label{tab:diverse_ablation_2}
\end{table}

\noindent\textbf{Generalizability of anchor-based sampling.}
In Table~\ref{tab:diverse_ablation_2}, we demonstrate that our anchor-based sampling is \textit{model-agnostic} and can be inserted as a \textit{plug-and-play} module into different motion predictors. Concretely, we apply our anchor-based sampling to the baseline method GSPS~\cite{mao2021generating}, which also achieves consistent improvements in every metric, with improvements in terms of diversity and multi-modal accuracy being particularly evident. For simplicity, this evaluation uses simple single-level anchors, but the improvements are pronounced. We would also like to emphasize that the total number of parameters in our \oursmodel{} predictor is only $\mathbf{22\%}$ of that in GSPS.

\subsection{Qualitative Results of Diverse Prediction}\label{sec:qual}

\begin{figure}[t]
\centering
\includegraphics[width=\textwidth]{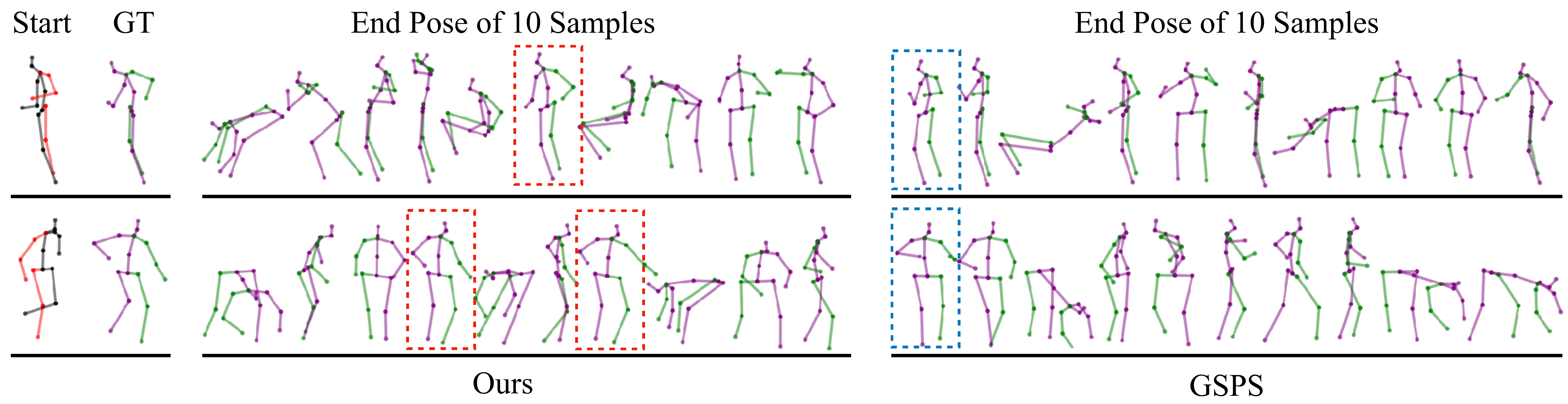}
\caption{\textbf{Visualization of end poses on Human3.6M.} We show the historical poses in red and black skeletons, and the predicted end poses with purple and green. As highlighted by the red and blue dashed boxes, the best predictions of our method are closer to the ground truth than the state-of-the-art baseline GSPS~\cite{mao2021generating}}
\label{fig:vis}
\end{figure}

\begin{figure}[t]
\centering
\includegraphics[width=\textwidth]{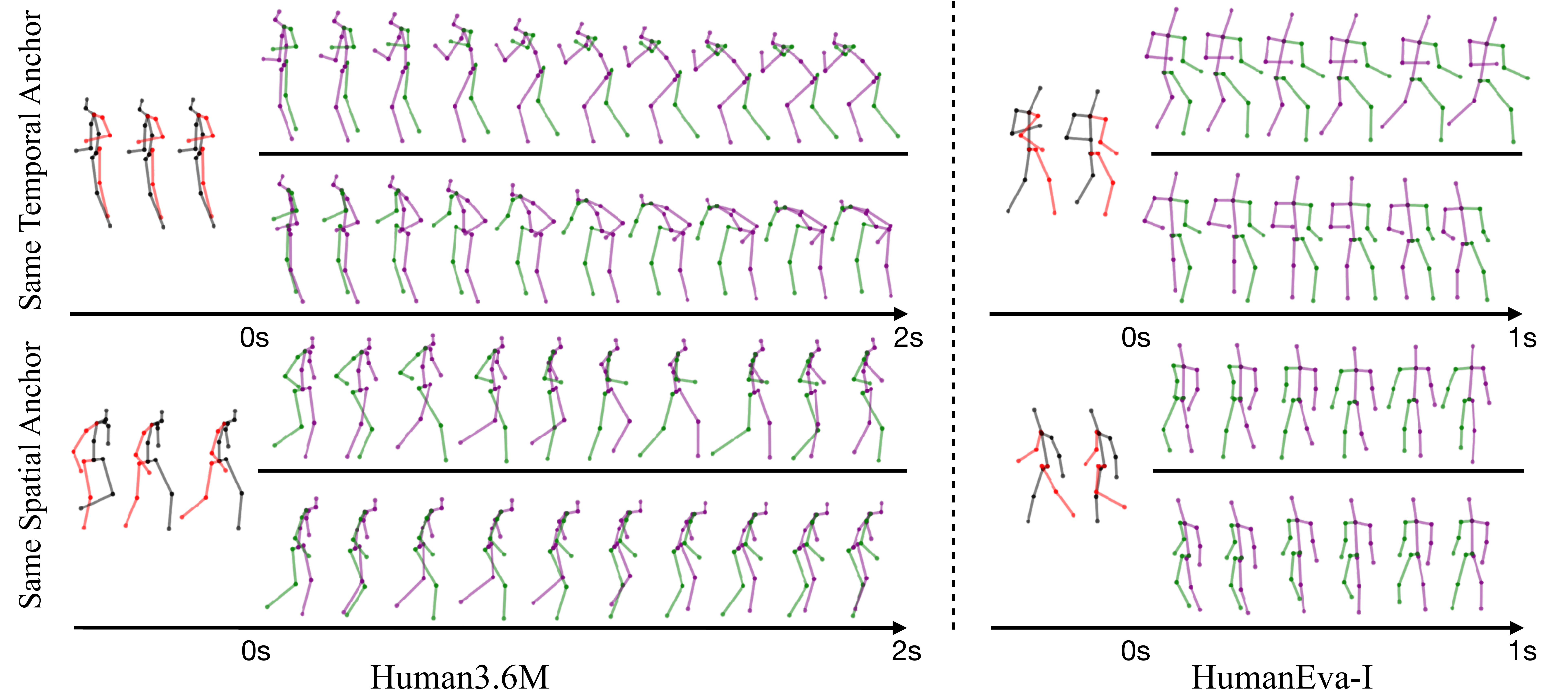}
\caption{\textbf{Visualization of controllable motion prediction} on Human3.6M and HumanEva-I. We control different trends and speeds of motion by controlling spatial and temporal anchors. For example, the third and fourth rows have similar motion trends, but the motion in the third row is faster}
\label{fig:vis_control}
\end{figure}

\begin{figure}[t]
\centering
\includegraphics[width=\textwidth]{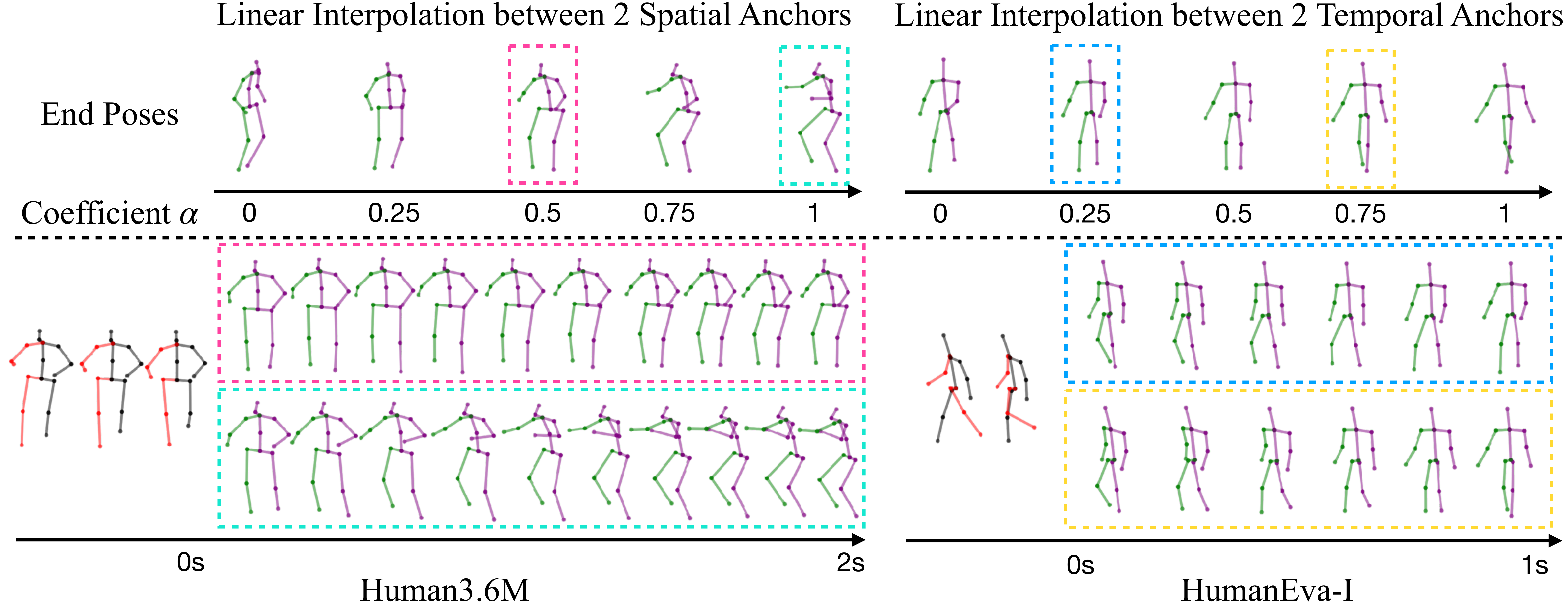}
\caption{\textbf{Linear interpolation of anchors.} We seamlessly control different trends and speeds of future motions by linear interpolation of spatial and temporal anchors. Specifically, given two anchors $\mathbf a_1$ and $\mathbf a_2$ and a coefficient $\alpha$, we produce predictions from the interpolated anchor formulated as $(1-\alpha)\mathbf a_1 + \alpha \mathbf a_2$}
\label{fig:vis_control_interpolate}
\end{figure}

We visualize the start pose, the end pose of the ground truth future motions, and the end pose of 10 motion samples in Fig.~\ref{fig:vis}. 
Qualitative comparisons support the ADE results in Table~\ref{tab:diverse_quan} that our best predicted samples are closer to the ground truth. In Fig.~\ref{fig:vis_control} and Fig.~\ref{fig:vis_control_interpolate}, we provide the predicted sequences sampled every ten frames. As mentioned, our spatial-temporal anchors provide a new form of control over spatial-temporal aspects. With the same temporal anchor, the motion frequencies are similar, but the motion patterns are different. Conversely, if we control the spatial anchors to be the same, the motion trends are similar, but the speed might be different. We show a smooth control through the linear interpolation of spatial-temporal anchors in Fig.~\ref{fig:vis_control_interpolate}. The new interpolated anchors produce some interesting and valid pose sequences. And smooth changes in spatial trend and temporal velocity can be observed.

\subsection{Effectiveness on Deterministic Prediction}
Our model can be easily extended to the deterministic prediction by specifying $K=1$. Without diverse sampling, we retrain two deterministic prediction model variants: \oursmodel-Short dedicated to the short-term prediction and \oursmodel-Long for long-term prediction.
We use different settings for deterministic prediction, following existing work~\cite{wei2019motion,Sofianos_2021_ICCV,Dang_2021_ICCV} and for fair comparisons. Here, we use the 22-joint skeleton representations on Human3.6M. Given a 400 ms past motion sequence, the model generates a 400 ms motion for short-term prediction and a 1000 ms motion for long-term prediction.
We use five subjects (S1, S6, S7, S8 and S9) for training and a subject (S5) for testing.
We compare our two model variants with state-of-the-art deterministic prediction baselines: \textbf{LTD}~\cite{wei2019motion}, \textbf{STS-GCN}~\cite{Sofianos_2021_ICCV}, and \textbf{MSR-GCN}~\cite{Dang_2021_ICCV}. We evaluate this by reporting Mean Per Joint Position Error (MPJPE)~\cite{h36m_pami} in \textit{millimeter} at each time step, defined as $\frac{1}{V} \sum_{i=1}^V\|\widehat{\mathbf{y}}_{t}[i] - \mathbf{y}_{t}[i]\|_2$, where $\mathbf{\widehat{y}}_{t}[i]$ and $\mathbf{y}_{t}[i]$ are produced and the ground truth 3D positions of the $i$-th joint at time $t$.
Table~\ref{tab:deter_ablation} includes short-term (80$\sim$400 ms) and long-term (560$\sim$1000 ms) comparisons, showing that our models outperform baseline models in both short-term and long-term horizons.
Additional experimental results and implementation details of two deterministic prediction models are provided in Sec.~\ref{sec:deterministic_supp} of the supplementary material.

\begin{table}[t]
	\caption{\textbf{Quantitative results} on Human3.6M for $K=1$. Both our long-term and short-term deterministic models significantly outperform all deterministic baselines}
	\centering
	\resizebox{.6\linewidth}{!}{
		\begin{tabular}{ccccccccccc}
		\noalign{\smallskip}
		\noalign{\smallskip}
			\toprule
			 \multirow{2}{*}{Method} & & \multicolumn{4}{c}{Short-term prediction} & & \multicolumn{4}{c}{Long-term prediction} \\
			  \cmidrule{3-6} \cmidrule{8-11}
			 & &\textit{80} & \textit{160} & \textit{320} & \textit{400}   & & \textit{560} & \textit{720} & \textit{880} & \textit{1000} \\ \midrule\midrule
			 LTD~\cite{wei2019motion} & &11.2 & 23.3 & 47.9 & 59.3 & & 79.9 & 94.3 & 106.1 & 113.3 \\
			 STS-GCN~\cite{Sofianos_2021_ICCV} & &13.5 & 27.7 & 54.4 & 65.8 & & 85.0 & 98.3 & 108.9 & 117.0 \\
			 MSR-GCN~\cite{Dang_2021_ICCV} & &11.3 & 24.3 & 50.8 & 61.9 & & 80.0 & 93.8 & 105.5 & 112.9 \\\midrule
			 \oursmodel-Short (Ours) & &\textbf{9.7} & \textbf{21.2} & \textbf{44.5} & \textbf{55.5} &  & 77.1 & 91.1 & 102.6 & 110.1 \\
			 \oursmodel-Long (Ours) & &10.0 & 21.8 & 45.7 & 56.9 &  & \textbf{75.8} & \textbf{89.3} &  \textbf{100.8} & \textbf{108.4} \\
			\bottomrule
		\end{tabular}
	}
	\label{tab:deter_ablation}
\end{table}
\section{Conclusion}
In this paper, we present a simple yet effective approach, \oursanchor, to predict multiple plausible and diverse future motions. And our spatial-temporal anchors enable novel controllable motion prediction. To incorporate our spatial-temporal anchors, we propose a novel motion predictor \oursmodel. Extensive experiments on Human3.6M and HumanEva-I show the state-of-the-art performance of our unified approach for both diverse and deterministic motion predictions. In the future, we will consider human-scene interaction and investigate the integration of our predictor into human-robot interaction systems.

\noindent\small{\textbf{Acknowledgement.} This work was supported in part by NSF Grant 2106825, the Jump ARCHES endowment through the Health Care Engineering Systems Center, the New Frontiers Initiative, the National Center for Supercomputing Applications (NCSA) at the University of Illinois at Urbana-Champaign through the NCSA Fellows program, and the IBM-Illinois Discovery Accelerator Institute.}

\clearpage

\bibliographystyle{splncs04}
\bibliography{main}

\newpage
{\noindent \large \bf {Supplementary Material}}\\
\setcounter{table}{0}
\setcounter{figure}{0}
\renewcommand\thesection{\Alph{section}}
\renewcommand\thetable{\Roman{table}}
\renewcommand\thefigure{\Roman{figure}}

\appendix
{\noindent In this supplementary material, we first provide a visualization video for additional qualitative comparisons. Please refer to the video and Sec.~\ref{sec:video} for details. In Sec.~\ref{sec:adom}, we demonstrate the additional information of our approach including the network architecture, and implementation details for both stochastic and deterministic prediction. In Sec.~\ref{sec: ar}, for completeness, we provide additional qualitative results and comparisons, including a user study, and quantitative analysis for both stochastic and deterministic predictions on Human3.6M and HumanEva-I.}
\section{Visualization Video}
\label{sec:video}
We include a video at \url{https://sirui-xu.github.io/STARS/images/demo.mp4}, to provide more comprehensive visualizations of 3D human motion prediction. These visualizations show that indeed our approach produces more diverse sequences, which we attribute to our \oursanchor{} strategy, where anchors can explicitly locate diverse modes. We further show the visualization of controllable motion prediction and illustrate the motion variation at both spatial and temporal levels, suggesting our novel manipulation of future motion in the {\em native space} and {\em time}.

\section{Additional Details of Methodology} \label{sec:adom}

\subsection{Multi-Level Spatial-Temporal Anchor-Based Sampling}\label{sec:shmp_supp}
\noindent{\bf Architecture.}
Here we detail our \ours{} by formulating the incorporation of backbone and anchor-based sampling, as illustrated in Fig.~\ref{fig:deterministic2} of the main paper. For instance, we sample $\mathbf z \in p(\mathbf z)$ and select $i$-th spatial anchor $\mathbf a_i^{s} \in \mathbb{R}^{M\times V\times C^{(l)}}$ and $j$-th temporal anchor $\mathbf a_j^{t} \in \mathbb{R}^{M\times V\times C^{(l)}}$ at each level, where $(i, j)$ is the 2D spatial-temporal anchor index corresponding to the 1D index $k$.

\noindent(1) The 4th layer in Fig.~\ref{fig:deterministic2} of the main paper is denoted as
\begin{align}\label{eq:1_supp}
    \mathbf a_i^{s_1} \in \mathcal{A}_s^{(1)}, \mathbf a_j^{t_1} \in \mathcal{A}_t^{(1)}, \ \mathbf{H}^{(4)}_k=\sigma(\mathbf {Adj}_s^{(3)}\mathbf{Adj}_f^{(3)}(\mathbf{H}^{(3)} + \mathbf a_i^{s_1} + \mathbf a_j^{t_1})\mathbf W^{(3)}).
\end{align}

\noindent(2) The 5th layer in Fig.~\ref{fig:deterministic2} of the main paper is denoted as
\begin{align}\label{eq:3}
    \mathbf z \sim p(\mathbf z), \ \mathbf{H}^{(5)}_k=\sigma((\mathbf M_s \odot \mathbf {Adj}_s^{(4)})\mathbf {Adj}_f^{(4)}[\mathbf{H}_k^{(4)}: \mathbf z]\mathbf W^{(4)}),
\end{align}
where the 5th layer is the pruned layer, and $\mathbf M_s$ is the predefined mask used for spatial interaction pruning.

\noindent(3) The 6th layer in Fig.~\ref{fig:deterministic2} of the main paper is denoted as
\begin{align}\label{eq:2}
\mathbf a_i^{s_2} \in \mathcal{A}_s^{(2)}, \mathbf a_j^{t_2} \in \mathcal{A}_t^{(2)}, \ \mathbf{H}^{(6)}_k=\sigma(\mathbf {Adj}_s^{(5)}\mathbf{Adj}_f^{(5)}(\mathbf{H}_k^{(5)} + \mathbf a_i^{s_2} + \mathbf a_j^{t_2})\mathbf W^{(5)}).
\end{align}

\noindent{\bf Training.}
Recall that in Sec.~\ref{sec:method} of the main paper, we divide the loss functions into the following three categories: (1) reconstruction losses, including reconstruction error and multi-modal reconstruction error; (2) diversity promoting loss; (3) motion constraint losses, including history reconstruction error, pose prior, limb loss, and angle loss.
Here, we provide detailed formulations of these loss functions.

\noindent(1) Reconstruction error, encouraging the best prediction close to the ground truth, thus prompting the corresponding anchor capture a mode, denoted as
\begin{align}
    \mathcal {L}_r = \min_k \|\mathbf{\widehat Y}_k - \mathbf{Y}\|^2.
\end{align}
(2) Multi-modal reconstruction error~\cite{mao2021generating}, which encourages predictions to cover multi-modal ground truth and thus promotes anchors to capture more modes, is denoted as
\begin{align}
    \mathcal{L}_{mm} = \frac{1}{N}\sum_{n=1}^N\min_k\|\mathbf{\widehat{Y}}_k - \mathbf{Y}_n\|^2.
\end{align}

The multi-modal ground truth~\cite{Yuan2020Diverse} is defined as $\{\mathbf Y_n\}_{n=1}^N$, representing the possible future motions in multiple modes. Specifically, given a threshold $\epsilon$, we cluster the future motions with similar start pose, as $\{\mathbf Y_n\}_{n=1}^N = \{\mathbf Y_n|\|\mathbf X_n[T_h] - \mathbf{X}[T_h]\| \leq \epsilon \}$, where $\mathbf X_n$ is the historical pose sequence of $\mathbf Y_n$.

\noindent(3) Historical reconstruction error~\cite{wei2019motion} alleviates the discontinuity between prediction and history by bringing the recovered historical motion $\mathbf {\widehat X}_k$ close to the past sequence of ground truth $\mathbf {X}$. Recall that our model recovers the past motion via Inverse DCT (IDCT) as $\mathbf{\widehat X}_k=(\mathbf{C^\mathsf{T}}\widetilde{\mathbf {Y}}_k)_{1:T_h}$. We denote this loss as
\begin{align}
\mathcal {L}_h = \frac{1}{K}\sum_{k=1}^K \|\mathbf {\widehat X}_k - \mathbf{X}\|^2.
\end{align}

\noindent(4) Diversity-promoting loss~\cite{yuan2020dlow}, which explicitly promotes pairwise distances of predictions to ensure that the anchors do not collapse to the same, is defined as
\begin{align}
    \mathcal{L}_d = \frac{2}{K(K-1)}\sum_{j=1}^K\sum_{k=j+1}^K\exp(-\frac{\|\mathbf{\widehat{Y}}_j - \mathbf{\widehat{Y}}_k \|_1}{\alpha}).
\end{align}
(5) Pose prior, using a pretrained normalizing flow $p_{nf}$ to measure the likelihood of the generated human poses $\mathbf{\widehat Y}_k$. We use this module to constrain that the generated poses have a high probability in $p_{nf}$,
\begin{align}
    \mathcal{L}_{nf} = -\sum_{k=1}^K\log p_{nf}(\mathbf{\widehat Y}_k).
\end{align}
(6) Limb loss, constraining the limb length to be consistent with the ground truth, is denoted as
\begin{align}
    \mathcal {L}_l = \frac{1}{K}\sum_{k=1}^K \|\mathbf {\widehat L}_k - \mathbf{L}\|^2,
\end{align} where the limb length is defined as the distance between two physically connected joints, and the vector $\mathbf {\widehat L}_k$ includes limb lengths of all poses in $\mathbf {\widehat X}_k$.

\noindent(7) Angle loss constrains the angles of the human skeleton to be in some valid ranges.
Please refer to~\cite{mao2021generating} for more details on the pose prior, limb loss, and angle loss.

\noindent{\bf Additional Implementation Details.}
First, the number of channels of the 8 STGCN layers $C^{(l)}$ starts from $C^{(0)}=3$, then $128, 64, 128, 64, 128, 64,$ $128$, and finally $C^{(8)}=3$. Accordingly, we insert 128-dimensional anchors in the fourth and sixth layers and use a 64-dimensional random noise in the fifth layer. Our code is based on PyTorch~\cite{paszke2019pytorch}, and we use ADAM~\cite{kingma2014autoencoding} to train the model. The learning rate is set to $0.001$ and decayed after the $100$ epochs as
\begin{align}
    lr = 0.001\times (1.0-\frac{\max(0, epoch-100)}{400}).
\end{align}

For Human3.6M, the weight of each loss term $(\lambda_r, \lambda_{mm}, \lambda_{h}, \lambda_d, \lambda_{nf}, \lambda_{l}, \lambda_{a})$ is set to $(2, 1, 50, 160, 0.01, 500, 100)$. And the first $20$ DCT coefficients are used.

For HumanEva-I, the weight of each loss term $(\lambda_r, \lambda_{mm}, \lambda_{h}, \lambda_d, \lambda_{nf}, \lambda_{l}, \lambda_{a})$ is set to $(2, 1, 10, 32, 0.002, 50, 10)$. Only the first $8$ DCT coefficients are used.

We set the distance threshold $\epsilon$ for generating the multi-modal ground truth mentioned above to $0.5$ on both datasets.

\noindent {\bf Limitation.}
One potential limitation of our \oursanchor{} is that the number of spatial and temporal anchors is decided manually via cross-validation. We conduct several ablation studies in Fig.~\ref{fig:quan_num} of the main paper to investigate the impact of anchor number. A better strategy might be learning the appropriate number directly from the data.

\begin{figure}[t]
\centering
\includegraphics[width=0.90\textwidth]{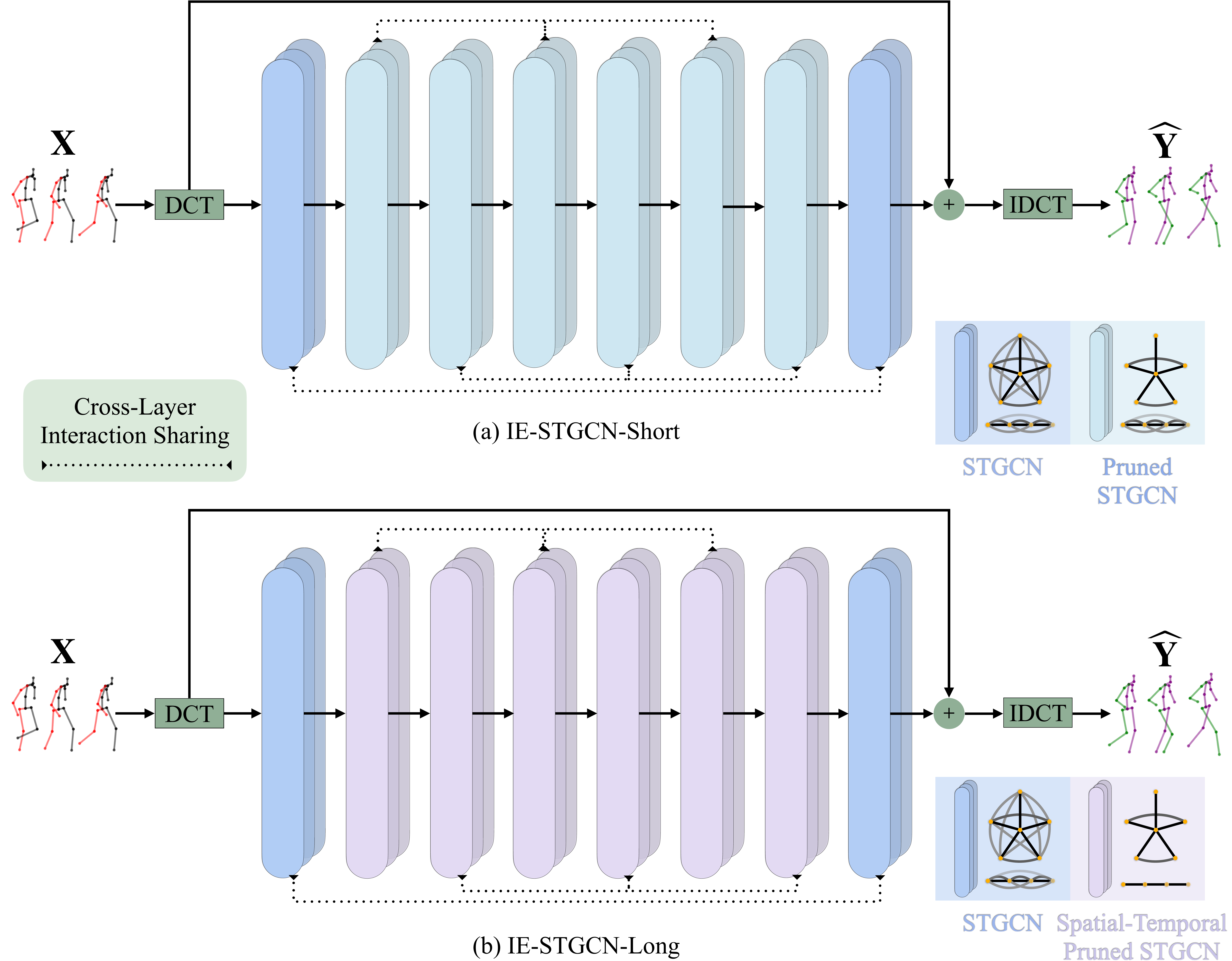}
\caption{\textbf{Overview of deterministic models.} For \oursmodel-Short, we set all middle layers as pruned STGCN layers, while we further apply temporal pruning to intermediate pruned layers for \oursmodel-Long}
\label{fig:deterministic2_supp}
\vspace{-1em}
\end{figure}

\subsection{Deterministic Human Motion Prediction}\label{sec:deterministic_supp}
\begin{table*}
	\caption{\textbf{Quantitative results} on Human3.6M and HumanEva-I for $K=50$. Our model significantly outperforms two additional baselines in all metrics. The results of two baselines are directly reported from \cite{tang2021probabilistic,SalzmannPavoneEtAl2022}}
	\footnotesize
	\centering
	\renewcommand{\arraystretch}{1.2}
\renewcommand{\tabcolsep}{1.2mm}
	\resizebox{\textwidth}{!}{
		\begin{tabular}{@{\hskip 0mm}c@{\hspace{5mm}}c@{\hspace{2mm}}c@{\hspace{2mm}}c@{\hspace{2mm}}c@{\hspace{2mm}}c@{\hspace{2mm}}c@{\hspace{2mm}}c@{\hspace{2mm}}c@{\hspace{2mm}}c@{\hspace{2mm}}c@{\hspace{2mm}}c@{\hskip 0mm}}
			\toprule
			\multirow{2}{*}{Method}& \multicolumn{5}{c}{Human3.6M~\cite{h36m_pami}} & & \multicolumn{5}{c}{HumanEva-I~\cite{Sigal:IJCV:10b}} \\ \cmidrule{2-6} \cmidrule{8-12}
			 & APD $\uparrow$ & ADE $\downarrow$ & FDE $\downarrow$ & MMADE $\downarrow$ & MMFDE $\downarrow$ & & APD $\uparrow$ & ADE $\downarrow$ & FDE $\downarrow$ & MMADE $\downarrow$ & MMFDE $\downarrow$ \\ \midrule
            ProTran~\cite{tang2021probabilistic} & / & 0.381 & 0.491 & / & / & & / & 0.258 & 0.255 & / & / \\
            Motron~\cite{SalzmannPavoneEtAl2022} & 7.168 & 0.375 & 0.488 & / & / & & / & / & / & / & /\\\hline
			\oursanchor{} (Ours) & \textbf{15.884} & \textbf{0.358} & \textbf{0.445} & \textbf{0.442} & \textbf{0.471} & & \textbf{6.031} & \textbf{0.217} & \textbf{0.241} & \textbf{0.328} & \textbf{0.321}\\
			\bottomrule
		\end{tabular}
	}
	\label{tab:diverse_quan_2}
\vspace{-1em}
\end{table*}
\noindent{\bf Architecture.} We illustrate our architecture for deterministic prediction \textit{without} \oursanchor{}, as shown in Fig.~\ref{fig:deterministic2_supp}. Note that there are only $4$ pruned STGCN layers for the stochastic prediction model. Here, we design all middle layers to be the pruned layers from layer $2$ to layer $7$. For \oursmodel-Long, we further apply temporal interaction pruning to these middle layers.

\noindent{\bf Temporal Interaction Pruning.} %
We emphasize the locality of frequency components by leveraging a temporal mask $\mathbf M_f$ to the frequency adjacency matrix of the spatial-temporal graph,
\begin{align}
    \mathbf {\widehat{Adj}}_f^{(l)} = \mathbf M_f \odot \mathbf {Adj}_f^{(l)}, \ \mathbf M_f[i][j] = \begin{cases} 1, & \mbox{for }|f_i - f_j| = 1, v_i= v_j \\ 0, & \mbox{otherwise}. \end{cases}
\end{align}

\noindent{\bf Training.}
We retain only two loss terms from Sec.~\ref{sec:shmp_supp}, \ie the reconstruction error, and the history reconstruction error. We denote the unique output as $\mathbf{\widehat Y}$, and the recovered historical motion as $\mathbf{\widehat X}$. These two loss terms are adapted as follows.

\setlength{\tabcolsep}{4pt}
\begin{table}[t]
\centering
\caption{\textbf{Quantitative results} of short-term prediction on Human3.6M. We compare our \oursmodel-Short with deterministic prediction baselines. We report the MPJPE error of 3D joint positions in \textit{millimeter}. Our model outperforms all baselines
}
\renewcommand{\arraystretch}{1.2}
\renewcommand{\tabcolsep}{1.2mm}
\resizebox{.7\textwidth}{!}{
\begin{tabular}{c@{\hspace{5mm}}cccc@{\hspace{5mm}}cccc@{\hspace{5mm}}cccc@{\hspace{5mm}}cccc}
				\noalign{\smallskip}
		\noalign{\smallskip}
\toprule
\multicolumn{1}{c}{Actions} &
  \multicolumn{4}{c}{Walking} &
  \multicolumn{4}{c}{Eating} &
  \multicolumn{4}{c}{Smoking} &
  \multicolumn{4}{c}{Discussion} \\ \hline
\multicolumn{1}{c}{\textit{msec}} &
  {\textit{80}} &
  {\textit{160}} &
  {\textit{320}} &
  {\textit{400}} & 
  {\textit{80}} &
  {\textit{160}} &
  {\textit{320}} &
  {\textit{400}} &
  {\textit{80}} &
  {\textit{160}} &
  {\textit{320}} &
  {\textit{400}} &
  {\textit{80}} &
  {\textit{160}} &
  {\textit{320}} &
  {\textit{400}} \\ \midrule

LTD \cite{wei2019motion} &
  11.1 &
  21.2 &
  37.4 &
  43.6 &
  7.3 &
  15.1 &
  29.6 &
  36.8 &
  7.1 &
  14.9 &
  29.5 &
  36.2 &
  11.5 &
  25.5 &
  55.8 &
  69.2 \\

STS-GCN \cite{Sofianos_2021_ICCV} &
  14.5 &
  26.0 &
  43.8 &
  51.0 &
  9.2 &
  18.0 &
  35.5 &
  43.3 &
  9.5 &
  18.1 &
  34.9 &
  42.3 &
  13.3 &
  27.9 &
  57.5 &
  71.9 \\
MSR-GCN \cite{Dang_2021_ICCV} &
  10.8 &
  20.9 &
  36.9 &
  42.4 &
  6.9 &
  14.6 &
  29.0 &
  36.0 &
  7.5 &
  15.4 &
  30.6 &
  37.5 &
  10.4 &
  23.5 &
  51.9 &
  65.0 \\\midrule
\multicolumn{1}{c}{\oursmodel-Short} &
  \textbf{10.0} &
  \textbf{19.4} &
  \textbf{35.1} &
  \textbf{41.6} &
  \textbf{6.3} &
  \textbf{13.8} &
  \textbf{28.3} &
  \textbf{35.9} &
  \textbf{6.4} &
  \textbf{13.1} &
  \textbf{25.8} &
  \textbf{32.3} &
  \textbf{9.0} &
  \textbf{19.9} &
  \textbf{42.8} &
  \textbf{55.2} \\ \bottomrule
\end{tabular}}

\resizebox{\textwidth}{!}{
\begin{tabular}{c@{\hspace{5mm}}cccc@{\hspace{5mm}}cccc@{\hspace{5mm}}cccc@{\hspace{5mm}}cccc@{\hspace{5mm}}cccc@{\hspace{5mm}}cccc}
\toprule
\multicolumn{1}{c}{Actions} &
  \multicolumn{4}{c}{Directions} &
  \multicolumn{4}{c}{Greeting} &
  \multicolumn{4}{c}{Phoning} &
  \multicolumn{4}{c}{Posing} &
  \multicolumn{4}{c}{Purchases} &
  \multicolumn{4}{c}{Sitting} \\ \hline
\multicolumn{1}{c}{\textit{msec}} &
  {\textit{80}} &
  {\textit{160}} &
  {\textit{320}} &
  {\textit{400}} &
  {\textit{80}} &
  {\textit{160}} &
  {\textit{320}} &
  {\textit{400}} &
  {\textit{80}} &
  {\textit{160}} &
  {\textit{320}} &
  {\textit{400}} &
  {\textit{80}} &
  {\textit{160}} &
  {\textit{320}} &
  {\textit{400}} &
  {\textit{80}} &
  {\textit{160}} &
  {\textit{320}} &
  {\textit{400}} &
  {\textit{80}} &
  {\textit{160}} &
  {\textit{320}} &
  {\textit{400}} \\ \midrule

LTD \cite{wei2019motion} &
  8.3 &
  19.1 &
  43.5 &
  \textbf{54.5} &
  14.6 &
  30.8 &
  64.5 &
  79.9 &
  9.0 &
  18.6 &
  39.3 &
  49.2 &
  11.2 &
  25.8 &
  59.6 &
  76.5 &
  13.7 &
  30.2 &
  63.4 &
  77.7 &
  9.8 &
  20.3 &
  44.1 &
  55.7 \\

STS-GCN \cite{Sofianos_2021_ICCV} &
  10.2 &
  23.0 &
  50.6 &
  63.2 &
  17.0 &
  36.6 &
  72.5 &
  86.4 &
  11.0 &
  21.9 &
  44.0 &
  53.8 &
  13.7 &
  30.4 &
  67.3 &
  84.7 &
  16.3 &
  35.9 &
  70.5 &
  83.1 &
  11.9 &
  23.8 &
  49.3 &
  60.8 \\
MSR-GCN \cite{Dang_2021_ICCV} &
  7.7 &
  18.9 &
  44.7 &
  56.2 &
  15.1 &
  33.1 &
  70.9 &
  85.4 &
  9.1 &
  18.9 &
  39.9 &
  50.0 &
  10.3 &
  24.6 &
  59.2 &
  75.9 &
  13.3 &
  30.1 &
  63.6 &
  77.8 &
  9.8 &
  20.6 &
  44.2 &
  55.5 \\
\midrule
\multicolumn{1}{c}{\oursmodel-Short} &
  \textbf{7.1} &
  \textbf{17.8} &
  \textbf{43.1} &
  54.7 &
  \textbf{12.8} &
  \textbf{28.5} &
  \textbf{61.1} &
  \textbf{75.9} &
  \textbf{8.2} &
  \textbf{17.5} &
  \textbf{37.3} &
  \textbf{47.3} &
  \textbf{8.3} &
  \textbf{18.9} &
  \textbf{43.6} &
  \textbf{57.7} &
  \textbf{12.2} &
  \textbf{28.3} &
  \textbf{60.0} &
  \textbf{74.1} &
  \textbf{8.9} &
  \textbf{19.4} &
  \textbf{42.9} &
  \textbf{54.4} \\ \bottomrule
\end{tabular}}

\resizebox{\textwidth}{!}{
\begin{tabular}{c@{\hspace{5mm}}cccc@{\hspace{5mm}}cccc@{\hspace{5mm}}cccc@{\hspace{5mm}}cccc@{\hspace{5mm}}cccc@{\hspace{5mm}}cccc}
\toprule
\multicolumn{1}{c}{Actions} &
  \multicolumn{4}{c}{Sitting Down} &
  \multicolumn{4}{c}{Taking Photo} &
  \multicolumn{4}{c}{Waiting} &
  \multicolumn{4}{c}{Walking Dog} &
  \multicolumn{4}{c}{Walking Together} &
  \multicolumn{4}{c}{\textbf{Average}} \\ \hline
\multicolumn{1}{c}{\textit{msec}} &
  {\textit{80}} &
  {\textit{160}} &
  {\textit{320}} &
  {\textit{400}} &
  {\textit{80}} &
  {\textit{160}} &
  {\textit{320}} &
  {\textit{400}} &
  {\textit{80}} &
  {\textit{160}} &
  {\textit{320}} &
  {\textit{400}} &
  {\textit{80}} &
  {\textit{160}} &
  {\textit{320}} &
  {\textit{400}} &
  {\textit{80}} &
  {\textit{160}} &
  {\textit{320}} &
  {\textit{400}} &
  {\textit{80}} &
  {\textit{160}} &
  {\textit{320}} &
  {\textit{400}} \\ \midrule

LTD \cite{wei2019motion} &
  14.8 &
  \textbf{29.5} &
  \textbf{57.2} &
  \textbf{71.2} &
  9.1 &
  19.1 &
  41.1 & 
  51.8 &
  9.4 &
  19.7 &
  43.2 &
  54.6 &
  21.1 &
  41.2 &
  75.1 &
  88.8 &
  9.6 &
  19.2 &
  36.0 &
  43.1 &
  11.2 &
  23.3 &
  47.9 &
  59.3 \\

STS-GCN \cite{Sofianos_2021_ICCV} &
  18.2 &
  37.2 &
  66.2 &
  79.4 &
  10.8 &
  22.3 &
  47.7 &
  59.4 &
  11.7 &
  24.0 &
  49.6 &
  62.0 &
  24.3 &
  48.0 &
  85.1 &
  97.3 &
  11.7 &
  22.7 &
  41.7 &
  49.1 &
  13.5 &
  27.7 &
  54.4 &
  65.8 \\
MSR-GCN \cite{Dang_2021_ICCV} &
  15.4 &
  32.0 &
  60.7 &
  73.8 &
  8.9 &
  19.5 &
  43.1 &
  54.4 &
  10.4 &
  22.4 &
  50.7 &
  62.4 &
  24.9 &
  51.5 &
  100.3 &
  112.9 &
  9.2 &
  18.7 &
  35.7 &
  43.2 &
  11.3 &
  24.3 &
  50.8 &
  61.9 \\\midrule
\multicolumn{1}{c}{\oursmodel-Short} &
  \textbf{14.6} &
  30.8 &
  60.2 &
  72.8 &
  \textbf{7.9} &
  \textbf{17.8} &
  \textbf{39.7} &
  \textbf{50.3} &
  \textbf{7.9} &
  \textbf{17.6} &
  \textbf{40.2} &
  \textbf{51.6} &
  \textbf{18.4} &
  \textbf{38.2} &
  \textbf{74.4} &
  \textbf{87.9} &
  \textbf{8.3} &
  \textbf{17.2} &
  \textbf{33.3} &
  \textbf{41.2} &
  \textbf{9.7} &
  \textbf{21.2} &
  \textbf{44.5} &
  \textbf{55.5} \\ \bottomrule
\end{tabular}}
\label{tab:short_mm}
\vspace{-1em}
\end{table}
\noindent{\bf }(1) Reconstruction error:
\begin{align}
    \mathcal {L}_r = \min_k \|\mathbf{\widehat Y}_k - \mathbf{Y}\|^2 = \|\mathbf{\widehat Y} - \mathbf{Y}\|^2.
\end{align}
\begin{table}
  \caption{\textbf{User study on Human3.6M. }Pairwise human voting results for predicted motions. Under human evaluation, our predictions significantly outperform the baseline in terms of diversity, considering the motion fidelity}
  \centering
  \resizebox{0.4\textwidth}{!}{
  \begin{tabular}{@{\hskip 0mm}c@{\hspace{5mm}}c@{\hspace{2mm}}c@{\hspace{2mm}}c@{\hspace{2mm}}c@{\hspace{2mm}}c@{\hspace{2mm}}c@{\hspace{2mm}}c@{\hskip 0mm}}
  				\noalign{\smallskip}
		\noalign{\smallskip}
    \toprule
    \multirow{2}{*}{Model pair} & & \multicolumn{2}{c}{Motion diversity} \\ \cmidrule{3-4}
     & &  Ours & GSPS\\
    \midrule
    \oursanchor{} (Ours) vs.  && n/a & 62.1\% \\
    GSPS vs.  && 37.9\% & n/a \\
    \bottomrule
  \end{tabular}}
  \label{tab:user}
  \vspace{-1em}
\end{table}

\noindent{\bf }(2) History reconstruction error:
\begin{align}
    \mathcal {L}_h = \frac{1}{K}\sum_{k=1}^K \|\mathbf {\widehat X}_k - \mathbf{X}\|^2 = \|\mathbf {\widehat X} - \mathbf{X}\|^2.
\end{align}

\noindent{\bf Implementation Details.}
We adopt the same number of channels as demonstrated in Sec.~\ref{sec:shmp_supp}. We increase the batch size to $256$ and train the model for only $50$ epochs. The learning rate is set to $0.01$ and decays by a factor of $0.1$ every $5$ epochs after the $20$th epoch. For \oursmodel-Short, we use the first 20 DCT coefficients. For \oursmodel-Long, we use the first 35 DCT coefficients.

\setlength{\tabcolsep}{4pt}
\begin{table}[t]
\centering
\caption{\textbf{Quantitative results} of long-term prediction on Human3.6M. We compare our \oursmodel-Long with deterministic prediction baselines. We report the MPJPE error of 3D joint positions in \textit{millimeter}. Our model outperforms all baselines 
}
\renewcommand{\arraystretch}{1.2}
\renewcommand{\tabcolsep}{1.2mm}
\resizebox{0.7\textwidth}{!}{
\begin{tabular}{c@{\hspace{5mm}}cccc@{\hspace{5mm}}cccc@{\hspace{5mm}}cccc@{\hspace{5mm}}cccc}
				\noalign{\smallskip}
		\noalign{\smallskip}
\toprule
\multicolumn{1}{c}{Actions} &
  \multicolumn{4}{c}{Walking} &
  \multicolumn{4}{c}{Eating} &
  \multicolumn{4}{c}{Smoking} &
  \multicolumn{4}{c}{Discussion} \\ \hline
\multicolumn{1}{c}{\textit{msec}} &
  {\textit{560}} &
  {\textit{720}} &
  {\textit{880}} &
  {\textit{1000}} &
  {\textit{560}} &
  {\textit{720}} &
  {\textit{880}} &
  {\textit{1000}} &
  {\textit{560}} &
  {\textit{720}} &
  {\textit{880}} &
  {\textit{1000}} &
  {\textit{560}} &
  {\textit{720}} &
  {\textit{880}} &
  {\textit{1000}} \\ \cline{1-17}

LTD \cite{wei2019motion}& 52.3 & 55.7 & 58.1 & \textbf{59.2} & \textbf{49.9} & \textbf{60.9} & 69.2 & 74.2 & 50.1 & 59.8 & 67.4 & 72.1 & 90.7 & 105.4 & 114.9 & 120.4 \\

STS-GCN \cite{Sofianos_2021_ICCV} & 60.3 & 64.6 & 65.9 & 70.2 & 57.2 & 68.3 & 75.5 & 82.6 & 54.2 & 63.8 & 70.8 & 76.1 & 91.8 & 105.2 & 113.8 & 118.9 \\
MSR-GCN \cite{Dang_2021_ICCV}& 53.3 & 55.4 & 58.1 & 63.7 & 50.8 & 61.4 & 69.7 & 75.4 & 50.5 & 59.5 & 67.1 & 72.1 & 87.0 & 101.9 & 111.4 & 116.8 \\\cline{1-17}
\multicolumn{1}{c}{\oursmodel-Long} &
  \textbf{49.3} &
  \textbf{53.5} &
  \textbf{57.4} &
  61.1 &
  50.2 &
  61.1 &
  \textbf{69.1} &
  \textbf{74.1} &
  \textbf{44.2} &
  \textbf{51.8} &
  \textbf{59.0} &
  \textbf{64.3} &
  \textbf{74.0} &
  \textbf{85.1} &
  \textbf{94.1} &
  \textbf{100.4} \\ \bottomrule
\end{tabular}}

\resizebox{\textwidth}{!}{%
\begin{tabular}{c@{\hspace{5mm}}cccc@{\hspace{5mm}}cccc@{\hspace{5mm}}cccc@{\hspace{5mm}}cccc@{\hspace{5mm}}cccc@{\hspace{5mm}}cccc}
\toprule
\multicolumn{1}{c}{Actions} &
  \multicolumn{4}{c}{Directions} &
  \multicolumn{4}{c}{Greeting} &
  \multicolumn{4}{c}{Phoning} &
  \multicolumn{4}{c}{Posing} &
  \multicolumn{4}{c}{Purchases} &
  \multicolumn{4}{c}{Sitting} \\ \hline
\multicolumn{1}{c}{\textit{msec}} &
  {\textit{560}} &
  {\textit{720}} &
  {\textit{880}} &
  {\textit{1000}} &
  {\textit{560}} &
  {\textit{720}} &
  {\textit{880}} &
  {\textit{1000}} &
  {\textit{560}} &
  {\textit{720}} &
  {\textit{880}} &
  {\textit{1000}} &
  {\textit{560}} &
  {\textit{720}} &
  {\textit{880}} &
  {\textit{1000}} &
  {\textit{560}} &
  {\textit{720}} &
  {\textit{880}} &
  {\textit{1000}} &
  {\textit{560}} &
  {\textit{720}} &
  {\textit{880}} &
  {\textit{1000}} \\ \midrule

LTD \cite{wei2019motion} &
  76.0 &
  91.2 &
  103.0 &
  108.8 &
  105.0 &
  120.6 &
  133.2 &
  139.4 &
  67.9 &
  \textbf{82.2} &
  95.0 &
  \textbf{103.3} &
  111.2 &
  137.3 &
  159.2 &
  172.8 &
  100.3 &
  115.2 &
  127.7 &
  135.5 &
  79.5 &
  98.4 &
  113.5 &
  122.8 \\

STS-GCN \cite{Sofianos_2021_ICCV} &
  79.5 &
  92.9 &
  102.2 &
  109.6 &
  111.2 &
  122.4 &
  131.8 &
  136.1 &
  72.5 &
  87.9 &
  99.7 &
  108.3 &
  115.8 &
  142.4 &
  161.7 &
  178.4 &
  104.6 &
  119.4 &
  132.7 &
  141.0&
  82.0 &
  97.6 &
  110.9 &
  121.4 \\
MSR-GCN \cite{Dang_2021_ICCV}&
  \textbf{75.8} &
  \textbf{89.9} &
  \textbf{100.5} &
  \textbf{105.9} &
  106.3 &
  120.0 &
  131.5 &
  136.3 &
  67.9 &
  82.5 &
  95.8 &
  104.7 &
  112.5 &
  140.1 &
  162.8 &
  176.5 &
  \textbf{99.2} &
  \textbf{114.0} &
  \textbf{126.9} &
  \textbf{134.4} &
  77.6 &
  94.0 &
  107.7 &
  115.9 \\\midrule
\multicolumn{1}{c}{\oursmodel-Long} &
  76.5 &
  91.6 &
  102.4 &
  107.6 &
  \textbf{101.2} &
  \textbf{116.5} &
  \textbf{129.5} &
  \textbf{135.8} &
  \textbf{67.2} &
  \textbf{82.1} &
  95.1 &
  103.8 &
  \textbf{79.7} &
  \textbf{97.2} &
  \textbf{113.9} &
  \textbf{129.3} &
  100.3 &
  117.5 &
  131.1 &
  139.4 &
  \textbf{77.5} &
  95.0 &
  109.1 &
  117.6 \\ \bottomrule
\end{tabular}}
\resizebox{\textwidth}{!}{%
\begin{tabular}{c@{\hspace{5mm}}cccc@{\hspace{5mm}}cccc@{\hspace{5mm}}cccc@{\hspace{5mm}}cccc@{\hspace{5mm}}cccc@{\hspace{5mm}}cccc}
\toprule
\multicolumn{1}{c}{Actions} &
  \multicolumn{4}{c}{Sitting Down} &
  \multicolumn{4}{c}{Taking Photo} &
  \multicolumn{4}{c}{Waiting} &
  \multicolumn{4}{c}{Walking Dog} &
  \multicolumn{4}{c}{Walking Together} &
  \multicolumn{4}{c}{\textbf{Average}} \\ \hline
\multicolumn{1}{c}{\textit{msec}} &
  {\textit{560}} &
  {\textit{720}} &
  {\textit{880}} &
  {\textit{1000}} &
  {\textit{560}} &
  {\textit{720}} &
  {\textit{880}} &
  {\textit{1000}} &
  {\textit{560}} &
  {\textit{720}} &
  {\textit{880}} &
  {\textit{1000}} &
  {\textit{560}} &
  {\textit{720}} &
  {\textit{880}} &
  {\textit{1000}} &
  {\textit{560}} &
  {\textit{720}} &
  {\textit{880}} &
  {\textit{1000}} &
  {\textit{560}} &
  {\textit{720}} &
  {\textit{880}} &
  {\textit{1000}} \\ \midrule

LTD \cite{wei2019motion}&
  \textbf{98.2} & \textbf{119.1} & \textbf{136.1} & 147.1 & \textbf{76.8} & \textbf{95.0} & \textbf{110.3} & \textbf{120.4} & 76.8 & 91.0 & 102.3 & 109.5 & 108.3 & 121.2 & 135.8 & 146.3 & 56.3 & 61.9 & 65.5 & 68.2 & 79.9 & 94.3 & 106.1 & 113.3\\

STS-GCN \cite{Sofianos_2021_ICCV} &
  104.1 &
  121.4 &
  137.6 &
  148.4 &
  81.2 &
  99.6 &
  111.6 &
  126.3 &
  80.3 &
  95.0 &
  105.9 &
  113.6 &
  119.0 &
  129.0 &
  143.9 &
  151.5 &
  61.9 &
  65.4 &
  69.1 &
  72.5 &
  85.0 &
  98.3 &
  108.9 &
  117.0 \\
  
MSR-GCN \cite{Dang_2021_ICCV}&
  102.4 &
  122.7 &
  139.6 &
  149.3 &
  77.7 &
  96.9 &
  112.3 &
  121.9 &
  74.8 &
  87.8 &
  98.2 &
  105.5 &
  \textbf{107.7} &
  \textbf{120.8} &
  \textbf{135.7} &
  \textbf{145.7} &
  56.2 &
  60.9 &
  65.0 &
  69.5 &
  80.0 &
  93.8 &
  105.5 &
  112.9 \\
\midrule
\multicolumn{1}{c}{\oursmodel-Long} &
  100.9 &
  120.7 &
  136.6 &
  \textbf{146.8} &
  78.1 &
  96.9 &
  112.1 &
  122.1 &
  \textbf{71.7} &
  \textbf{85.0} &
  \textbf{96.0} &
  \textbf{103.6} &
  111.9 &
  126.3 &
  142.9 &
  153.1 &
  \textbf{54.1} &
  \textbf{59.5} &
  \textbf{63.5} &
  \textbf{67.5} &
  \textbf{75.8} &
  \textbf{89.3} &
  \textbf{100.8} &
  \textbf{108.4} \\ \bottomrule
\end{tabular}}
\label{tab:long_mm}
\vspace{-1em}
\end{table}

\section{Additional Results} \label{sec: ar}
\subsection{Quantitative Results} \label{sec:quantitative}

\noindent{\bf Comparison with Additional Baselines.} Motron~\cite{SalzmannPavoneEtAl2022} provides a flexible output structure, which can produce deterministic predictions by weighting each mode by a confidence value. Another baseline is ProTran~\cite{tang2021probabilistic}, a deep probabilistic method combining transformer architectures and state space models. However, these two baselines do not provide results for all standard metrics in the literature. In Table~\ref{tab:diverse_quan_2}, we compare our method with them. For all metrics reported, our method still consistently outperforms baselines.

\subsection{Qualitative Results}
\noindent{\bf User study on Human3.6M.}
In Tables~\ref{tab:diverse_quan} and~\ref{tab:diverse_ablation} of the main paper, we compare our method with the state-of-the-art method GSPS~\cite{mao2021generating}. We measure the quality of the best predictions and overall diversity according to the metrics demonstrated in Sec~\ref{sec: metrics} of the main paper. Here, we conduct a user study to evaluate the diversity of predicted human motions under consideration of motion fidelity. The reason for this user study is that there may be unrealistic predictions, or outliers, that result in very large APD, but do not affect the ADE since ADE measures only the quality of the best prediction. We evaluate and rule out such ``cheating'' behavior through user studies.

\begin{figure}[t]

\centering
\includegraphics[width=\textwidth]{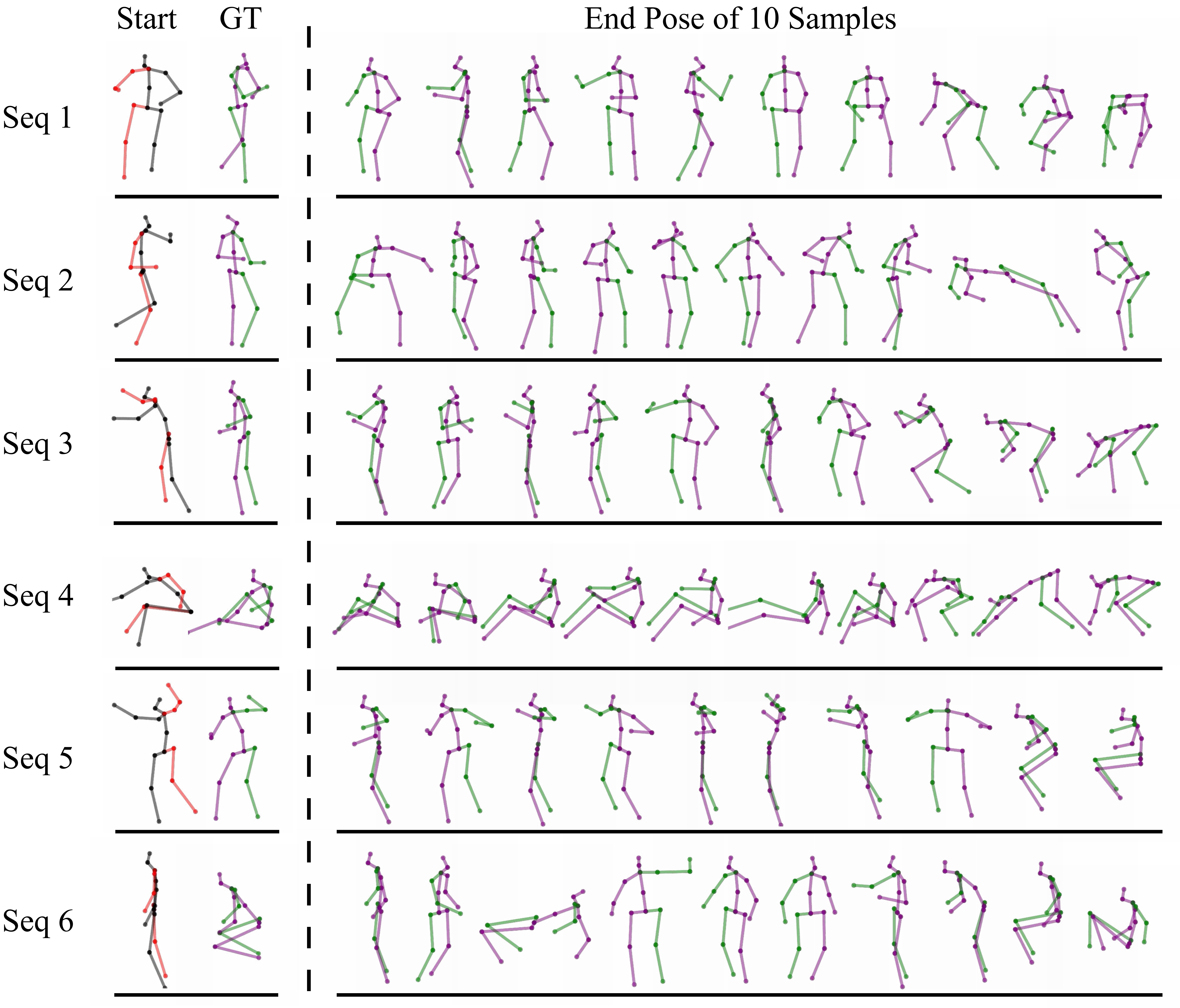}
\caption{\textbf{Additional visualization on Human3.6M.} We show the start pose, the end pose of ground truth future motion, and the end poses of ten samples predicted by our approach}
\label{fig:h36m}
\vspace{-1em}
\end{figure}

We conduct a {\em double-blind} user study. 
We randomly sample $20$ input sequences on Human3.6M. For GSPS, we randomly sample two predicted sequences. For our \ours{}, we randomly sample two predictions {\em generated by different spatial-temporal anchors}. We design pairwise evaluations. Considering one pair from ours and another one from GSPS, human judges are asked to determine which pair is more diverse given the action labels, taking into account motion fidelity.

From the results of the human evaluations in Table~\ref{tab:user}, our approach has a success rate of $62.1\%$ against GSPS, verifying that our method generates more diverse and valid motions than the baseline method.

\begin{figure}[t]
\centering
\includegraphics[width=\textwidth]{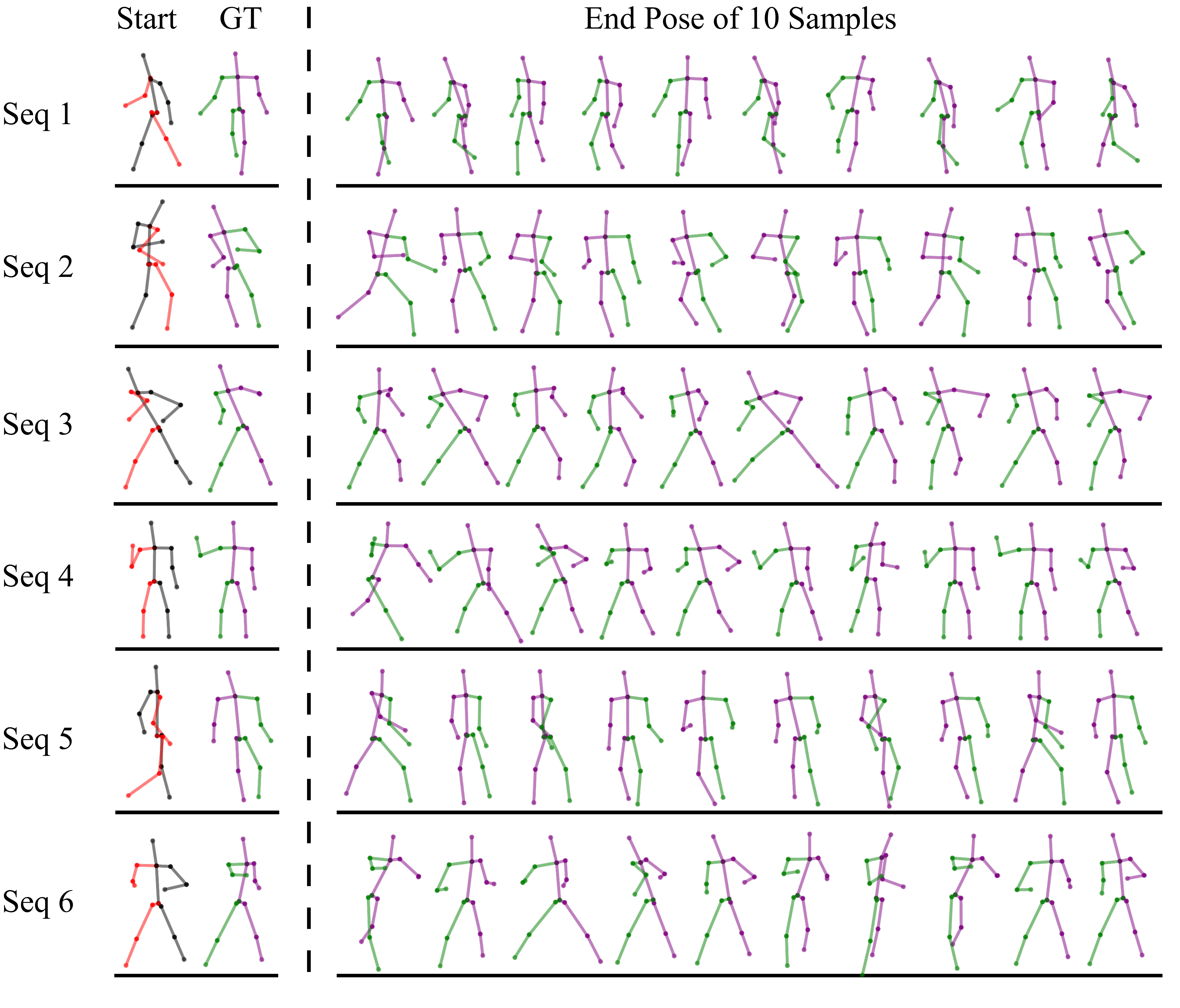}
\caption{\textbf{Additional visualization on HumanEva-I.} We show the start pose, the end pose of ground truth future motion, and the end poses of ten samples predicted by our approach}
\label{fig:eva}
\vspace{-1em}
\end{figure}

\noindent{\bf Additional Visualizations on Human3.6M and HumanEva-I. } In Fig.~\ref{fig:h36m} and Fig.~\ref{fig:eva}, we provide additional qualitative results by visualizing the end poses of $10$ samples.

\begin{figure}[t]
\centering
\includegraphics[width=\textwidth]{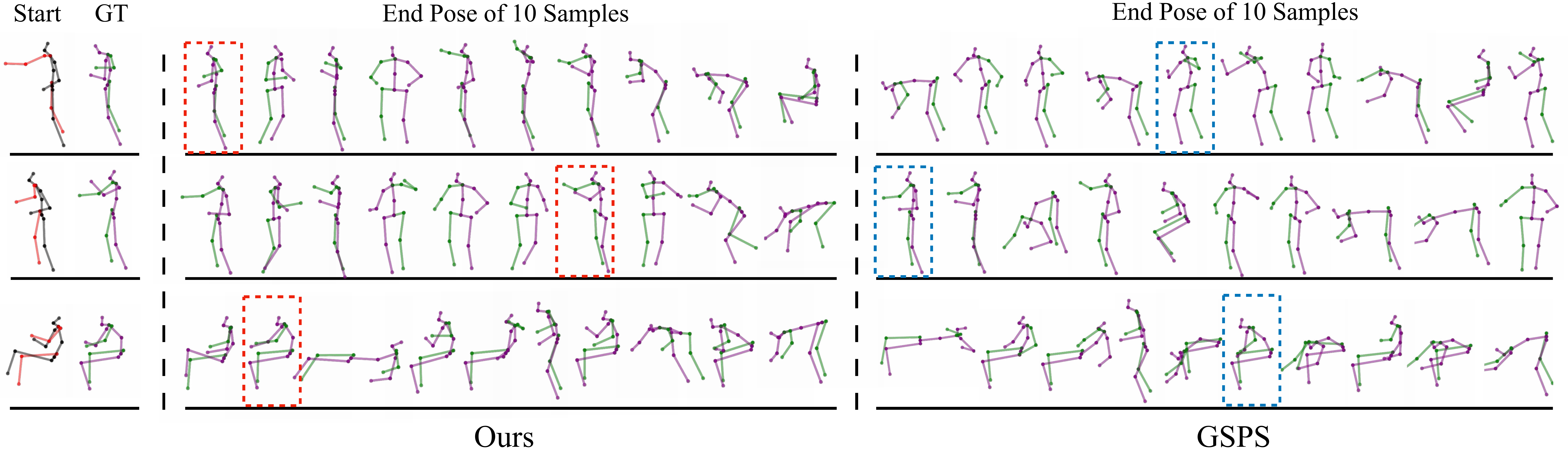}
\caption{\textbf{Additional comparisons with the baseline on Human3.6M.} As highlighted by the red and blue dashed boxes, the best predictions of our method are closer to the ground truth than the state-of-the-art baseline GSPS~\cite{mao2021generating}}
\label{fig:gsps}
\end{figure}

\begin{table}[t]

	\caption{Average MPJPE error in mm on Human3.6M, comparing different spatial-temporal enhancement techniques. ``TP" indicates whether we prune the temporal interaction. ``TS" means if we share the temporal adjacency matrix across some of the layers (See Fig.~\ref{fig:deterministic2_supp}). "TF" stands for start GCN and end GCN containing unpruned fully temporal interactions. ``SP", ``SS" and ``SF" have the similar meaning but apply to temporal interaction}
	\centering
	\resizebox{0.65\linewidth}{!}{
		\begin{tabular}{@{\hskip 0mm}cccccccccccccccccc@{\hskip 0mm}}
						\noalign{\smallskip}
		\noalign{\smallskip}
			\toprule
			\multirow{2}{*}{TP} & \multirow{2}{*}{TS} & \multirow{2}{*}{TF} & \multirow{2}{*}{SP} & \multirow{2}{*}{SS} & \multirow{2}{*}{SF} & \multicolumn{4}{c}{Short-term prediction} & & \multicolumn{4}{c}{Long-term prediction} \\ 
			\cmidrule{7-10} \cmidrule{12-15}
			 &&&&&& \textit{80} & \textit{160} & \textit{320} & \textit{400}   & & \textit{560} & \textit{720} & \textit{880} & \textit{1000} \\ \midrule
			 & & \checkmark &  &  & \checkmark & 10.1 & 21.7 & 44.9 & 55.8 & & 79.8 & 94.3 & 106.3 & 113.9 \\\midrule
		      & & \checkmark & \checkmark & \checkmark & \checkmark & 9.8 & 21.3 & 44.8 & 55.8 &  & 77.0 & 91.1 & 102.7 & 110.1  \\
			 \checkmark & & & \checkmark & \checkmark & \checkmark & 10.2 & 22.1 & 46.2 & 57.2 & & 77.1 & 90.8 & 101.9 & 109.6 \\
			 & \checkmark & \checkmark &\checkmark & \checkmark & \checkmark & \textbf{9.7} & \textbf{21.2} & \textbf{44.5} & \textbf{55.5} &  & 77.1 & 91.1 & 102.6 & 110.1 \\
			 \checkmark & \checkmark &  & \checkmark & \checkmark & \checkmark & 23.8 & 44.4 & 76.1 & 88.2 &  & 76.9 & 90.2 & 101.3 &109.2\\
			 \checkmark&  &\checkmark  & \checkmark & \checkmark & \checkmark & 10.0 & 21.8 & 45.7 & 56.9 &  & 75.8 & 89.3 &  100.8 & 108.4 \\
			 \checkmark & \checkmark & \checkmark & \checkmark & \checkmark & \checkmark & 10.0 & 21.8 & 45.7 & 56.9 &  & \textbf{75.7} & 89.4 & \textbf{100.8} & 108.5\\
			 \checkmark&  &\checkmark  & \checkmark & \checkmark & \checkmark & 10.0 & 21.8 & 45.7 & 56.9 &  & 75.8 & \textbf{89.3} &  \textbf{100.8} & \textbf{108.4} \\
			 \checkmark&  &\checkmark  & \checkmark & &\checkmark & 9.9 & 21.6 & 45.1 & 56.3 &  & \textbf{75.7} & 89.5 & \textbf{100.8} & \textbf{108.4} \\
			 \checkmark&  &\checkmark  & & \checkmark & \checkmark & 9.9 & 21.7 & 45.3 & 56.3 &  & 76.3 & 89.9 & 101.3 & 109.0 \\
			 \checkmark&  &\checkmark  & \checkmark & \checkmark & & 10.3& 23.2& 48.5 & 59.8 &  & 79.0 & 92.5 &  103.7 & 111.2\\
			\bottomrule
		\end{tabular}
	}

	\label{tab:ablation}

\end{table}

\noindent{\bf Additional Qualitative Comparisons with GSPS Baseline.} 
We provide comparisons with GSPS in Fig.~\ref{fig:gsps}, in addition to Fig.~\ref{fig:vis} of the main paper. Our model still successfully generates predictions that are closer to the ground truth.

\begin{figure}
\centering
\includegraphics[width=\textwidth]{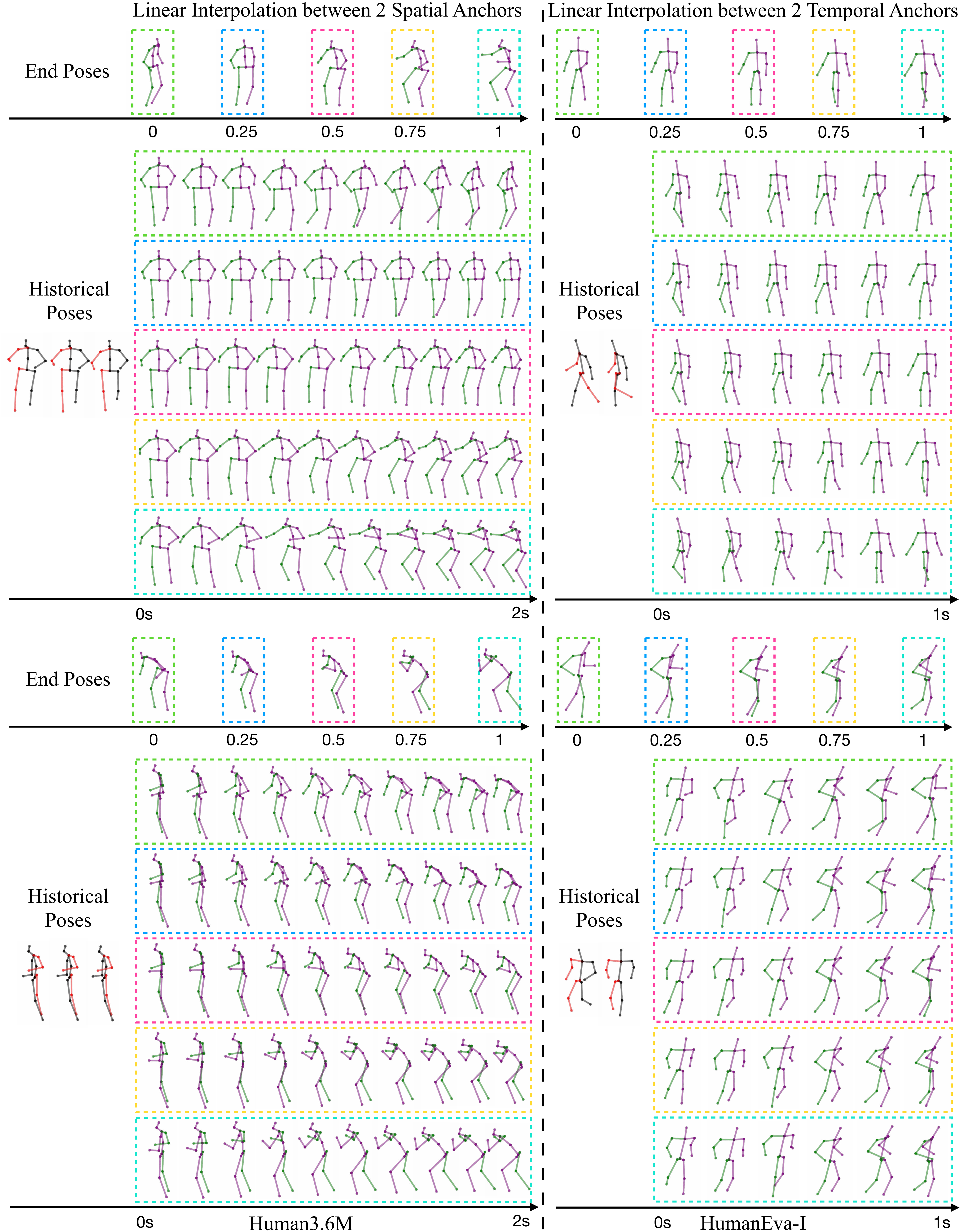}
\caption{\textbf{Linear interpolation of anchors. }We show additional visualizations on Human3.6M and HumanEva-I. We provide seamless control over directions and speeds of future motion by linear interpolation of the spatial and temporal anchors}
\label{fig:interpolate}
\end{figure}

\noindent{\bf Additional Visualizations on Controllable Motion Prediction.}
In addition to Fig.~\ref{fig:vis_control_interpolate} of the main paper, we present the visualization of the controllable motion prediction in Fig.~\ref{fig:interpolate}.

\subsection{Additional Results on Deterministic Prediction}
\noindent{\bf Effectiveness on Deterministic Prediction.} 
To evaluate the deterministic prediction, we randomly select $256$ sequences for each action category. We {\em re-evaluate} the pretrained models provided by baseline approaches~\cite{wei2019motion,Dang_2021_ICCV,Sofianos_2021_ICCV} with the same selection of test data.
Note that we re-evaluate STS-GCN using a standard evaluation metric for a fair comparison, reporting MPJPE in millimeters at each frame, rather than the average MPJPE over all frames as in~\cite{Sofianos_2021_ICCV}.

Here, we provide detailed results and comparisons across all actions. As shown in Table~\ref{tab:short_mm} and Table~\ref{tab:long_mm}, our model achieves state-of-the-art short- and long-term prediction performance. 

\noindent\textbf{Effectiveness of interaction enhancements.} We conduct an ablation study in Table~\ref{tab:ablation} to demonstrate the effectiveness of our proposed interaction enhancements in deterministic predictions. 
The results show that with proper use of interaction enhancements, we could improve the accuracy for both the short- and long-term deterministic predictions.
We observe that the spatial interaction enhancements are effective for both short-term and long-term horizon, while temporal enhancements are helpful only for long-term prediction. The reason may be that in the long-term prediction, we use more DCT coefficients, which may contain redundancies that need to be pruned.

\clearpage

\end{document}